%% file: main.tex
\documentclass[journal,final]{IEEEtran}

\usepackage{amsmath}
\usepackage{bbm}
\usepackage{amssymb}
\usepackage{amsthm}	
\usepackage[usenames, dvipsnames]{xcolor}
\usepackage{algorithm}
\usepackage[noend]{algpseudocode}

\usepackage[noadjust]{cite}

\usepackage{enumerate}
\usepackage{graphicx}

\makeatletter
\let\MYcaption\@makecaption
\makeatother

\usepackage[font=footnotesize]{subcaption}

\makeatletter
\let\@makecaption\MYcaption
\makeatother

\usepackage{subcaption}
\usepackage{mathtools}

\usepackage{float}
\usepackage{flushend}
\usepackage{longtable}

\usepackage{soul}
\usepackage{array}

\usepackage{tikz}
\usetikzlibrary{arrows,automata,positioning}

\usepackage{hyperref}

\input{symbols}
\begin{document}
%
\title{On Sensing, Agility, and Computation Requirements\\ for a Data-gathering Agile Robotic Vehicle}
%
%
%

\author{Fangchang Ma \qquad\qquad\qquad\qquad\qquad Sertac Karaman
\thanks{The authors are with the Department
of Aeronautics and Astronautics, Laboratory for Information and Decision Systems,
Massachusetts Institute of Technology, Cambridge,
MA, 02139}
}

\maketitle

\begin{abstract}

We consider a robotic vehicle tasked with gathering information by visiting a set of spatially-distributed data sources, the locations of which are not known {\em a priori}, but are discovered on the fly. We assume a first-order robot dynamics involving drift and that the locations of the data sources are Poisson-distributed. 
In this setting, we characterize the performance of the robot in terms of its sensing, agility, and computation capabilities. More specifically, the robot's performance is characterized in terms of its ability to sense the target locations from a distance, to maneuver quickly, and to perform computations for inference and planning. We also characterize the performance of the robot in terms of the amount and distribution of information that can be acquired at each data source.
The following are among our theoretical results: the distribution of the amount of information among the target locations immensely impacts the requirements for sensing targets from a distance; performance increases with increasing maneuvering capability, but with diminishing returns; and the computation requirements increase more rapidly for planning as opposed to inference, with both increasing sensing range and maneuvering ability.
We provide computational experiments to validate our theoretical results. 
Finally, we demonstrate that these results can be utilized in the co-design of sensing, actuation, and computation capabilities of mobile robotic systems for an information-gathering mission. 
Our proof techniques establish novel connections between the fundamental problems of robotic information-gathering and the last-passage percolation problem of statistical mechanics, which may be of interest on its own right. 

\end{abstract}


%
\IEEEpeerreviewmaketitle

\input{introduction}

\input{problem-formulation}

\input{preliminaries}

\input{analysis-continuous}
\input{experiments}


\input{discussion}


\section{Conclusion}
\label{sec:conclusion}
In this paper, we propose the maximum-reward motion problem for studying fundamental limits of data-gathering robots, given their sensing, actuation and computation constraints. We model the robot as a particle moving in a stochastic reward field and analyze its performance by using results from last-passage percolation problem in statistical mechanics. We verify our theoretical results in thorough simulation experiments. We also apply our results in the design of data-gathering vehicles as well as sensor selection for unattended ground sensors, providing insights for these problems.

\bibliographystyle{IEEEtran}
\bibliography{reference}


%

\appendices
\input{appendices/proof-lattice-limit-exists}
\input{appendices/proof-lattice-optimal-light-tailed}
\input{appendices/proof-lattice-exponential-geometric}
\input{appendices/proof-lattice-optimal-pareto}
\input{appendices/proof-lattice-sensing-light-tailed}
\input{appendices/proof-lattice-sensing-pareto}

\input{appendices/proof-cont-limit-exists}
\input{appendices/proof-cont-optimal-heavy-tailed}
\input{appendices/proof-cont-optimal-pareto}
\input{appendices/proof-cont-agility}

\end{document}

%% file: symbols.tex
\newcommand{\ie}{{\em i.e.}}
\newcommand{\eg}{{\em e.g.}}

\newcommand{\totalreward}{T}
\newcommand{\meanreward}{R}

\newcommand{\optimaltotalreward}{T^*_d}
\newcommand{\optimalmeanreward}{R^*_d}
\newcommand{\optimaltotalrewardtwo}{T^*_2}
\newcommand{\optimalmeanrewardtwo}{R^*_2}
\newcommand{\optimaltotalrewardlattice}[1]{T_{(#1)}}

\newcommand{\expectedoptimalmeanreward}{\mathbf{\meanreward}^*}
\newcommand{\optimalmeanrewardTruncated}{\mathbf{\meanreward}^{*(L)}_d}

\newcommand{\IMPtotalreward}{\totalreward_\mathrm{iterative}}
\newcommand{\IMPmeanreward}{\meanreward_\mathrm{iterative}}
\newcommand{\IMPmeanrewardTruncated}{\meanreward_\mathrm{iterative}^{(L)}}

\newcommand{\pathset}{\Pi}
\newcommand{\pathname}{\pi}

\newcommand{\bfx}{\mathbf{x}}
\newcommand{\poisson}{\mathbf{Pois}}
\newcommand{\rewardApproximate}{r^{(N)}}

\newcommand{\sensingrange}{S}
\newcommand{\agility}{\alpha}
\newcommand{\planningtime}{C_P}
\newcommand{\interencetime}{C_I}

\newcommand{\finaltime}{\tau}

\newcounter{constants}
\newcommand{\nextconstant}{\stepcounter{constants}c_\theconstants}
\newcommand{\thisconstant}{c_\theconstants}

\newcommand{\vertexnumber}[1]{v_{#1}}
\newcommand{\vertex}{\mathbf{v}}
\newcommand{\reals}{\mathbb{R}}
\newcommand{\E}{\mathbb{E}}
\newcommand{\naturals}{\mathbb{N}}
\newcommand{\EE}{\mathbb{E}}
\newcommand{\PP}{\mathbb{P}}

\DeclarePairedDelimiter\floor{\lfloor}{\rfloor}

\let\oldmarginpar\marginpar
\renewcommand\marginpar[1]{\oldmarginpar[\raggedleft\footnotesize #1]%
{\raggedright\footnotesize #1}}

\newcommand{\hide}[1]{ }

\newtheorem{theorem}{Theorem}

\newtheorem{lemma}{Lemma}
\newtheorem{proposition}{Proposition}

\newtheorem{definition}{Definition}

\newtheorem{remark}{Remark}

\newcommand{\myIncludeFourFigures}[4]
{
\newcommand{\minipageWidth}{0.24\linewidth}
\centering
\begin{minipage}{\textwidth}
\def\arraystretch{10}		
\setlength\tabcolsep{0.5mm}	
\scriptsize{
\begin{tabular}{cc}
\begin{minipage}{\minipageWidth}%
\centering%
\includegraphics[width=\linewidth]{#1} \\
(a)
\end{minipage}
&  
\begin{minipage}{\minipageWidth}%
\centering%
\includegraphics[width=\linewidth]{#2} \\
(b)
\end{minipage}
\\ 
\begin{minipage}{\minipageWidth}%
\centering%
\includegraphics[width=\linewidth]{#3} \\
(c)
\end{minipage}
&
\begin{minipage}{\minipageWidth}%
\centering%
\includegraphics[width=\linewidth]{#4} \\
(d)
\end{minipage}
\end{tabular}
}
\end{minipage}%
}

\newcommand{\myIncludeTwoFigures}[6]
{
\newcommand{\minipageWidth}{0.235\linewidth}
\centering
\begin{minipage}{\textwidth}
\def\arraystretch{10}		
\setlength\tabcolsep{1mm}	
\scriptsize{
\begin{tabular}{cc}
\begin{minipage}{\minipageWidth}%
\centering%
\includegraphics[width=\linewidth, trim={#3 #4 #5 #6}, clip]{#1} \\
(a)
\end{minipage}
&  
\begin{minipage}{\minipageWidth}%
\centering%
\includegraphics[width=\linewidth, trim={#3 #4 #5 #6}, clip]{#2} \\
(b)
\end{minipage}
\end{tabular}
}
\end{minipage}%
}

\newcolumntype{K}[1]{>{\centering\arraybackslash}m{#1}}
\newcommand{\specialcell}[2][c]{%
  \begin{tabular}[#1]{@{}K{5cm}@{}}#2\end{tabular}}

%% file: introduction.tex

\section{Introduction}
\label{section:introduction}
%
%
%
%

\begin{figure}[hbt]
\centering
\includegraphics[width=0.97\linewidth]{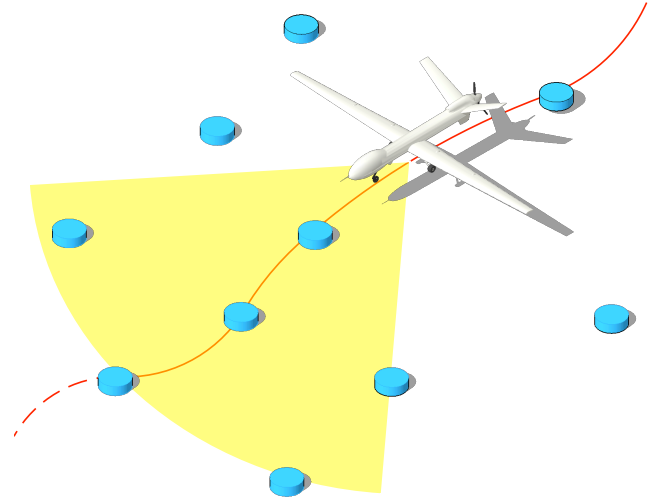}
\caption{An illustration of the vehicle navigating in a stochastic reward field. The blue cylinders represent the target locations. The yellow region represents the target-detection region attached to the vehicle. The locations of all targets in this range are known to the vehicle. By visiting these target locations, the vehicle can collect the reward assigned to them, as illustrated by the vehicle trajectory in red.}
\label{figure:introductory_illustration}
\end{figure}

The Unmanned Aerial Vehicle (UAV) technology thrived during the last decade.  Today, the diverse UAV market offers products with a wide range of size, power, speed, endurance, agility, and payload capacity properties, and with staggering granularity. 
The ability to produce and utilize UAVs with such diversity has made a number of new civilian applications commercially viable, particularly in the agriculture, security, humanitarian assistance, and disaster response domains, as evidenced by the large and increasing number of technology companies that aim to utilize UAVs in these domains. 


Developing the enabling hardware, algorithms, and software still remains one of the prominent challenges.
However, given the diversity of the UAV market today and the range of the potential applications, new research challenges emerge around the system design aspects, for instance, to answer the question: {\em How can we choose the best UAV system that can address a given UAV mission, with provable guarantees on performance?} 

For example, consider the persistent monitoring of an agricultural field. Suppose the UAVs are tasked with locating and picturing the potential anomalies, where the locations of anomalies are not known to the UAV {\em a priori}, but they are discovered on the fly. Once the location with a potential anomaly is detected from a distance, \eg, after that location enters the robot's vision range, the observing UAV may choose to fly over for a close-up picture to confirm the anomaly. Then, given the size of the field and how frequently the potential anomalies arise, how shall we choose the sensing, agility, and computation capabilities of the UAVs, in order to ensure that anomalies can be detected before they adversely affect the field? Note that the choice of these properties ultimately determines the the size, power, endurance, speed, and payload capacity of the UAV that shall be utilized in this mission. 

Currently, such design problems are addressed in an ad-hoc manner, for instance through guesswork, or at best through extensive simulation studies. Unfortunately, rigorous mathematical tools that provide valuable insight into these system design problems still remain largely unavailable, with the exception of a few very recent results which we discuss below.

In this paper, we focus on a problem that involves information gathering, which arises in a number of UAV applications. We consider the design of sensing, agility, and computation capabilities of a robotic vehicle that is tasked with gathering information from spatially-distributed data sources. 
Specifically, we prove a number of results that characterize the performance of the system in terms of {\em (i)} the distance from which the robot can locate the data sources, {\em (ii)} the agility of the robot, {\em (iii)} the onboard computational capabilities of the robot, and {\em (iv)} the amount and distribution of information at the field. These results characterize certain sensing, agility, and computation requirements for UAVs designed to execute a certain class data gathering missions. As a result, they provide insight into how to best select UAVs for the mission at hand. 

Our technical approach is to reduce a certain class of data gathering problems to a \emph{maximum reward collection problem}. First, we show that a discrete version of this maximum reward collection problem is closely related to a widely-studied model in statistical mechanics, namely the \emph{last-passage percolation} model. Then, we analyze this problem utilizing certain results available in the statistical mechanics literature. Finally, we extend our results in the discrete domain to continuous spaces. 

\subsection{Related Work} There are a number of relevant problems studied in the literature, including the traveling salesman problem, the vehicle routing problem, the orienteering problem, and the persistent monitoring problem. 
%

\subsubsection{Traveling Salesman Problem (TSP)} The TSP problem has been widely studied in the theoretical computer science and the operations research literature~\cite{bellmore1968traveling, lawler1985traveling, laporte1992traveling, applegate2011traveling}. Given a list of $n$ cities and the distances between every pair of them, the TSP problem is to find the shortest tour that visits each of the cities once before returning to the starting city. 
Although the problem is NP-hard in general, practical heuristics~\cite{lin1973effective}, polynomial-time approximations~\cite{rosenkrantz1974approximate}, branch-and-bound techniques~\cite{padberg1991branch} and learning algorithms~\cite{dorigo2014ant} have been developed. 
In relation to the present paper, Beardwood {\em et al.} \cite{beardwood1959shortest} consider the stochastic TSP problem where the $n$ points are independently and uniformly distributed over a bounded region. 
%
More recently, the traveling salesperson problem for the Dubins vehicle (DTSP), \ie, a non-holonomic vehicle that is constrained to move along planar paths of bounded curvature, was also studied~\cite{savla2008traveling, le2012dubins}. 

\subsubsection{Vehicle Routing Problem (VRP)} VRP is a generalization of the TSP problem~\cite{dantzig1959truck, christofides1976vehicle, toth2001vehicle, golden2008vehicle}. In the VRP, a fleet of vehicles, which start from a central depot, seek to service customers. 
Similar to TSP, a number of exact and approximate algorithms have been proposed for solving this problem~\cite{laporte1992vehicle, desrochers1992new, osman1993metastrategy, gendreau1994tabu, baldacci2012recent}. A version of the VRP problem for single vehicle is due to Wilson and Colvin~\cite{wilson1977computer}, where customer requests appear dynamically. Dynamic and stochastic routing problems are also studied as an extension of their deterministic counterparts~\cite{gendreau1996stochastic}. For example, Bertsimas and Ryzin~\cite{bertsimas1991stochastic} consider the the dynamic traveling repairman problem (DTRP), in which the demands for service arrive according to a Poisson process, 
and they provide a lower bound on the average total waiting time of demands. 

The underlying maximum-reward motion problem that we study also has a stochastic and dynamic nature, since the targets are stochastically placed and they are encountered in a dynamic manner as the targets get in the sensing range of the vehicle. 
However, the maximum-reward motion problem differs from the stochastic and dynamic versions of the TSP and the VRP in a number of ways. Most importantly, in the maximum-reward motion problem, the vehicle is governed by dynamics with drift and it is {\em not} constrained to visit all of the target locations. 
Moreover, the dynamic nature arises from the vehicle's motion, rather than as a spatio-temporal process that is independent of the motion of the vehicle. Also, the reward placed on the targets is an essential part of the problem. Since, TSP and VRP problems constrain the vehicle to visit all of the targets, placing reward on the targets does not change the solution.
Hence, we believe that the maximum-reward motion problem introduced in this paper is a fundamental stochastic dynamic decision-making problem on its own right. Its differentiating key attribute is the following: The drift affecting the vehicle does not allow visiting all of the targets, and the vehicle must choose targets based on the reward. 

\subsubsection{Orienteering Problem (OP)} The OP originates from orienteering, an outdoor sport where a set of locations are specified. Scores are associated with these locations, and competitors seek visit a subset of these locations within a fixed amount of time in order to maximize the total score. The orienteering problem is closely tied to the traveling salesman problems (TSP) and the vehicle routing problems (VRP), and has been studied extensively in the operations research community. Its variants are also known as the maximum collection problem~\cite{kataoka1988algorithm, butt1994heuristic}, selective traveling salesman problem~\cite{laporte1990selective}, traveling salesman problems with profits~\cite{feillet2005traveling}, and the bank robber problem~\cite{arkin1998resource}. This problem has been shown to be NP-hard~\cite{golden1987orienteering}, and the recent paper~\cite{vansteenwegen2011orienteering} provides a complete survey on the literature of the OP, including their variants, applications and solutions. Several heuristic methods have been proposed to tackle the OP since the early days, including deterministic and Monte-Carlo based heuristics~\cite{tsiligirides1984heuristic}, a so-called center-of-gravity heuristic~\cite{golden1987orienteering}, and many more~\cite{chao1996fast, golden1988multifaceted}. Other algorithms are also presented in the literature, including branch-and-cut algorithms~\cite{fischetti1998solving} for finding the optimal solution, optimization-base methods~\cite{leifer1994strong}, and search-based algorithms~\cite{geem2005harmony}. Researchers have also extended the OP to the case with multiple vehicles, also known as the Team Orienteering Problem~\cite{chao1996team, tang2005tabu, archetti2007metaheuristics, vansteenwegen2009iterated, vansteenwegen2009guided, labadie2012team}.

The difference between the orienteering problem and this maximum-reward problem is three-fold. Firstly, in the maximum-reward problem the motion of robot is governed by its dynamics (and drift).
Secondly, the OP problem is deterministic, and in comparison the maximum-reward problem has stochastic and dynamic nature arising from the robot's motion.
Finally, the OP problem considers a finite-distance path (or limited-time path); whilst in this maximum-reward problem, our analysis focuses on the scaling of the performance. 

\subsubsection{Persistent Monitoring} The data-gathering problem which we consider is also similar to the recently-proposed persistent monitoring problems that seek to generate an optimal control strategy for a team of agents to monitor a dynamically changing environment. There exist many different formulations and applications of the persistent monitoring problem~\cite{smith2011persistentAdapting, garg2014persistent}. For example, a persistent monitoring task with the objective to minimize an uncertainty metric is considered in~\cite{cassandras2011optimal, lin2013optimal, cassandras2013optimal}. A different formulation of the persistent monitoring problem is studied in~\cite{smith2011persistent, smith2012persistent}, where robots move along fixed paths in a changing environment. In \cite{alamdari2014persistent, alamdari2013min}, Alamdari {\it et. al} consider path planning problem for a robot to monitor a known set of features of interest, where the environment is represented as a weighted graph. In a more recent work~\cite{yu2014persistent}, Yu {\em et al.} focus on a stochastic model of occurrence of events, where the precise occurrence time is not known {\em a priori}.

The data-gathering problem which we study differs from the problems in these references with its stochastic and dynamic nature. Moreover, our analysis fundamentally differs from these references, since we focus mainly on the asymptotic performance analysis of an optimal algorithm for a fundamental problem, while the references develop complex algorithms, often based on mathematical programming.

The particular special case that we focus on is similar to the dynamic boundary guarding problem~\cite{smith2009dynamic, bopardikar2010dynamic} studied by Smith {\it et. al}, where the service robot they study is constrained on a line segment while our robot can move freely along the $x_2$ dimension so long as the speed is bounded.  
Our work differs in two key aspects. First, our problem definition is more general in a number of aspects. We consider problems where the agent can move freely, whereas the aforementioned references consider a vehicle confined to a box. We attach a reward to each target and aim to maximize total reward, while the said references references maximize the number of targets visited, which is recovered when all reward is equal in our problem. 
%
%
Second, our proof techniques utilize techniques from statistical mechanics, specifically the last-passage percolation problem, whereas they base their analysis on the methods of combinatorial optimization in stochastic domains. 
In fact, the generality we provide comes with the novel connections we establish with the last-passage percolation problem. 

\subsubsection{Cyber-Physical Systems} Our work is in the same domain of the recent work exploring the theory of cyber-physical systems design and co-design~\cite{matni:design05,censi16codesign,censi2017uncertainty}. This recent line of work aims to cast the design problem that arise in cyber-physical systems and robotics as optimization problems. Our work is complementary, and it can be considered in this emerging field of design.


It is worth noting that a dual of the maximum-reward problem has been studied~\cite{karaman2012high}. The authors investigate high-speed navigation through a randomly-generated obstacle field, where only the statistics of the obstacles are given. They show that with a simple dynamical model of the vehicle, the existence of an infinite collision-free trajectory through the environment exhibits a phase transition, \ie, there is an infinite collision-free trajectory almost surely when the speed is below a threshold and it will collide with some tree eventually otherwise. In~\cite{karaman2012highcdc}, they show that a planning algorithm based on state lattices can navigate the robot with limited sensing range. A similar problem is also studied in \cite{choudhurytheoretical}. 

\subsubsection{Percolation Theory and Queues in Tandem} Finally, our work is closely tied to the literature of percolation theory (specifically, last-passage percolation~\cite{rolla2008last,seppalainen1997increasing,seppalainen2009lecture,zeng2013directed,Grimmet:2002vk,Hambly:2007wm}) and related research on queues in tandem~\cite{baccelli2000asymptotic, glynn1991departures}. A complete survey on last-passage percolation model with general weights can be found in~\cite{martin2006last}. It can be shown that under certain technical assumptions the two models are equivalent, and results from the percolation theory have been applied in systems of queues in tandem. Recent work by Somanath {\em et al.} also applies similar models to control theory and robotics~\cite{somanath2014controlling}.

\subsection{The Maximum-reward Motion Problem}
As we briefly surveyed above, there are a number of problems that have been studied towards understanding missions involving autonomous vehicles in data gathering applications. However, we observe that there are no foundational problems that include sensing, agility, and computation properties at the same time in a stochastic environment. We close this gap by introducing a new foundational problem. Different from the existing literature, this problem involves a vehicle moving with drift in an environment where the tasks are distributed randomly and each target is associated with a random reward. The vehicle is endowed with a limited sensing range. In this foundational problem, the vehicle is tasked with collecting the maximum reward as it navigates through its environment. 
This foundational problem, which we call the {\em maximum-reward motion problem}, is illustrated in Figure~\ref{figure:introductory_illustration}.
We present this problem in detail in Section~\ref{section:problem:problem1}. We present an important special case in Section~\ref{section:problem:problem2}, which we believe is the simplest version of the problem that still maintains its core properties. 
In Sections \ref{section:preliminaries} and~\ref{section:analysis}, we analyze this new foundational problem. 

This foundational problem is connected with problems involving data gathering with agile robotic vehicles. We outline these connections briefly in Section~\ref{section:problem:data_gathering}, and then we devote Section~\ref{section:application} to describe this connection in detail. 

\subsection{Contributions}
A preliminary version of this paper appeared in the Workshop on Algorithmic Foundations of Robotics~\cite{ma2015maximum}, where we introduced some of the analysis for the discrete lattices presented in Section~\ref{section:preliminaries}. However, the other results in this paper, including all results in Section~\ref{section:analysis}, are new. The contribution of this paper is three-fold. First, we formulate the maximum-reward motion problem, which serves as a novel mathematical foundation for the analysis of a class of data-gathering problems in robotics. Second, we provide a rigorous analysis of the robot performance, given its sensing, actuation and computation capabilities. This is achieved by establishing connections with the last-passage percolation problem in statistical mechanics. Third, we apply our results to gain insights for the design of UAV systems.

Our thorough analysis reveals a number of insights, which are explained in detail in Section~\ref{section:application}. In particular, we find:
\begin{table*}[ht]
\centering
\captionof{table}{Summary of results} \label{tab:summary} 
    \begin{tabular}{ | K{5cm} | K{5cm} | K{5cm} |}
    \hline
    & \specialcell{Light-tailed distributed rewards\\(\eg, bounded, exponential, geometric, gaussian)}
    & \specialcell{Heavy-tailed distributed rewards\\(\eg, Pareto, Cauchy, Student's t distributions)}
    \\ \hline
    \specialcell{Optimal unit-distance mean reward $\optimalmeanrewardtwo(L)$\\(with unlimited sensing range)}
    & $\optimalmeanrewardtwo(L)$ converges to a finite constant $\optimalmeanrewardtwo$.
    & \specialcell{$\optimalmeanrewardtwo(L)$ grows unbounded, \ie, $\optimalmeanrewardtwo=\infty$. \\ For example, for Pareto distributions with parameter $\alpha$, $\optimalmeanrewardtwo(L) = O(L^{2/\alpha-1})$} 
    \\ \hline
    Required sensing range $\sensingrange$ for mission length $L$  
    & $\sensingrange=O(\log(L))$ 
    & $\sensingrange=O(L)$ \\ \hline
    Impact of robot agility $\agility$ on mean reward
    & \multicolumn{2}{ c | }{$\optimalmeanrewardtwo = O(\agility^{1/2})$ }  
    \\ \hline
    Computational requirement for motion planning
    & \multicolumn{2}{ c | }{$\planningtime=O\left(\lambda^2 \agility^2 \sensingrange^3 v \right)$}
    \\ \hline
    Computational requirement for inference tasks
    & \multicolumn{2}{ c | }{$\interencetime=O\left(\sqrt{\agility}\,v\right)$}
    \\ \hline
    \end{tabular}
\end{table*}
\begin{itemize}
\item
The vehicle can navigate almost optimally, \ie, as if it had infinite sensing range, even with very little sensing range (required sensing range scales only logarithmically with increasing mission length), when the value of information is bounded almost surely for each target location; on the contrary, when the value of information on each target is Pareto distributed (a heavy tailed distribution), the required sensing range that is almost as large as the mission length to perform optimally.
\item
The impact of agility on the performance of the agent can be completely characterized for the simple example that we consider. In our metrics, performance increases with increasing agility, but with diminishing returns. 
\item
The computation requirements can also be completely quantified in terns of the sensing range and the agility of the vehicle. We find that the computational effort devoted to planning increases faster than that devoted to inference, as various parameters increase.
\end{itemize}

A summary of results can be found in Table~\ref{tab:summary}.

\subsection{Organization}
We formalize the maximum-reward motion problem in Section~\ref{section:problem}. We introduce and analyze a discrete version of the problem in Section~\ref{section:preliminaries}. With the help of the results for the discrete case, we study the continuous problem in Section~\ref{section:analysis}. In Section~\ref{sec:experiments}, we provide the results of simulations that support our theoretical results. Finally in Section~\ref{section:application}, we present applications of maximum-reward motions to a sensor selection problem as well as a design problem that involves a network of UAVs and unattended ground sensors.

%% file: problem-formulation.tex

\section{Problem Formulation}
\label{section:problem}
This section is devoted to a formal definition of the problem. For this purpose, we first define the problem of collecting maximum reward in a stochastic reward field in its most general form. Second, we introduce an important special case, which this paper focuses on. Finally, as an instance of this problem, we introduce an inference problem involving mobile robotic vehicles tasked with data gathering.


\subsection{Problem 1: Maximum-reward Motion for Vehicles with Drift Operating in a Stochastic Environment}
\label{section:problem:problem1}

Consider a robotic vehicle navigating in a stochastic environment, where the locations of targets are distributed randomly and each target location is associated with a random reward value. The precise locations of all of the targets are unknown to the robot {\em a priori}. Instead, the vehicle discovers the target locations and the reward associated with the targets on the fly. To model this phenomenon, we consider a target-detection region attached to the vehicle.  When the targets get inside the detection region of the robot, the locations of the targets and the reward associated with them become known to the robot. The vehicle can then choose which locations to visit and collect the reward associated with these visited targets. 

Note that, when subject to differential constraints involving substantial drift, the vehicle must visit the most valuable targets in the direction of drift selectively, in order to maximize the total reward it collects. This often comes at the expense of skipping some of the target locations, for instance, those that are orthogonal to the drift direction. 
See Figure~\ref{figure:introductory_illustration}.


In this section, we present the reward collection problem in a general form. 
In the next section, we introduce a special case that captures all key aspects of the problem. This special case is also analytically tractable. In particular, we can derive the aforementioned fundamental limits for this special case. 

The online motion planning problem is formalized as follows in its most general form:

{\bf Dynamics:} 
Consider a mobile robotic vehicle that is governed by the following equations: 
\begin{align}
\begin{array}{l}
\dot{x}(t) = f(x(t), u(t)),\\
y(t) = g(x(t))
\end{array}
\label{equation:system}
\end{align}
where $x(t) \in X \subset \reals^n$ represents the state, $u(t) \in U \subset \reals^m$ represents the control input, $y(t) \in \reals^2$ is the position of the robot on the plane where the targets lie, $X$ is called the state space, and $U$ is called the control space. 
A state trajectory $x : [0,\finaltime] \to X$ is said to be a {\em dynamically-feasible state trajectory} and $y: [0,\finaltime] \to \reals^2$ is said to be a {\em dynamically-feasible output trajectory}, if there exists $u : [0,\finaltime] \to U$ such that $u$, $y$, and $x$ satisfy Equation~\eqref{equation:system} for all $t \in [0,\finaltime]$.

We are particularly interested in the case when the robot is subject to drift, for instance, when the robot can not come to a full stop instantly or can not even substantially slow down.%
\footnote{Let us note at this point that ``dynamical systems with drift'' can be defined precisely, for instance, through differential geometry~\cite{ivancevic2007applied}. However, we will not need such differential-geometric definitions in this paper, since we focus on a particular system with drift (introduced in Section~\ref{section:problem:problem2}), and we leave the generalization to other drift systems to future work.}
Examples include fixed-wing airplanes, racing cars, large submarines, and speed boats. In Section~\ref{section:problem:problem2}, we will present a dynamic system model, which we believe is the simplest model that captures this drift phenomenon.

{\bf Targets and reward:} The target locations and the reward associated with the targets are assumed to be generated by a stochastic marked point process.%
\footnote{Strictly speaking, stochastic point processes are formalized using counting measures~\cite{daley2007introduction}. For the sake of the simplicity of the presentation, we will avoid these measure-theoretic constructs, and instead we will use the simpler notation adopted by Stoyan et al.~\cite{chiu2013stochastic}.}
A marked point process is defined as a random, countably-infinite set of pairs $\{(p_i, m_i) : i \in \naturals\}$, where $p_i \in \reals^2$ is the location of point $i$ in the infinite plane and $m_i \in M$ is the mark associated with point $i$. We denote this random set by $\Psi$. With a slight abuse of notation, we denote by $\Psi(A)$ the number of points in a subset $A \in \reals^2$ of the infinite plane. Given a point $p$ of the point process, we denote its mark by $r(p)$. In our case, the locations $\{p_i\}$ of the points represent the locations of the targets, and the marks $\{m_i\}$ represent the reward associated with the targets. Hence, the mark set is the set of all non-negative real numbers, \ie, $M = \reals_{\ge 0}$. Following Stoyan et al.~\cite{chiu2013stochastic}, we make the following technical assumptions: {\em (i)} any bounded subset of the plane contains finitely many points, \ie, $\vert \Psi(A) \vert < \infty$ for all bounded measurable $A \subset \reals^2$; {\em (ii)} no two points are located at the same location, \ie, $p_i \neq p_j$ for all $i \neq j$, almost surely. 

{\bf Target-sensing region:} The locations of the targets and the reward associated with them is not known {\em a priori}, but is revealed to the robot in an online manner. This aspect of the problem is formalized as follows. Let ${\cal P}_\Psi(\cdot)$ denote the target-detection region of the robot that associates each state $z \in X$ of the robot with a region ${\cal P}_\Psi\left( z \right) \subset \reals^2$. When the robot is in state $z \in X$, it is able to observe only those targets that lie in the set ${\cal P}_\Psi\left( z \right)$. That is, $\{(p_i,m_i) \in \Psi : p_i \in {\cal P}_\Psi(z)\}$ is the set revealed to the robot when it is in state $z$. 

{\bf Task:} The robot is assigned the task of collecting maximum total reward, subject to all of the constraints outlined above. We formalize this objective of the problem as follows. 
Suppose the stochastic marked point process that represents the targets is defined on the probability space $(\Omega, {\cal F}, \PP)$, where $\Omega$ is the sample space, ${\cal F}$ is the $\sigma$-algebra, and $\PP$ is the probability measure. Define ${\cal F}_t$ as the $\sigma$-algebra generated by the random variables $\cup_{t' \in [0,t]} {\cal P}_\Psi(x(t'))$. 
A {\em feasible control policy} is a stochastic process $\mu = \{u(t) : t \in [0,\finaltime]\}$, such that $u(t)$ is defined on $(\Omega,{\cal F}_t, \PP)$, for all $t \ge 0$.\footnote{We omit some of the measure-theoretic details when defining the control policy. Our definition matches the definition of an adapted control policy introduced by Kushner~\cite{kushner1971introduction}.}
This definition implies that the control policy depends only on the locations and the reward of the targets that are detected by the robot up until time $t$, but not at times greater than $ t$. Note that the control policy may depend on the statistics of the stochastic marked point process, if any statistics are known {\em a priori}.

Given a control policy $\mu = \{u(t) : t \in [0,\finaltime]\}$, let us denote the collection of resulting state trajectories by $\{x_\mu(t) : t\in [0,\finaltime]\}$ and the collection of the output trajectories by $\{y_\mu(t) : t\in [0,\finaltime]\}$, which are stochastic processes defined on the same probability space as the control policies.

Let $Y(\mu;\psi)$ denote the set of targets visited by the robot under control policy $\mu$ and when the realization of the stochastic marked point process for the targets is $\psi$, \ie, 
$$
Y(\mu;\psi) = \big\{p \in \psi : y_\mu(t) = p \mbox{ for some }t \in [0,\finaltime]\big\}.
$$
With a slight abuse of notation, let $\totalreward (\mu;\psi)$ denote the total reward collected by the robot when it visits these targets, \ie, 
$$
\totalreward(\mu; \psi) = \sum_{p \in Y(\mu;\psi)} r(p).
$$

Then, the {\em maximum-reward motion problem} is to find a control policy $\mu$ such that the total reward $\totalreward(\mu;\psi)$ is maximized for all realizations of $\psi$ of the point process $\Psi$.


We stress that an algorithmic solution to this problem (\ie, computing such a policy) is often simple, particularly when the point process $\Psi$ is completely random (\eg, a Poisson process). Instead of designing new algorithms, we are more interested in analytically deriving the maximum reward that can be achieved by an optimal algorithm. Such analyses may allow robotics engineers to design robotic systems, \eg, by choosing the sensing, actuation, and computation capabilities of the robots, such that they best fit the application at hand. 

\vspace{-.1in}

\subsection{Problem 2: Maximum-reward Motion with Lipschitz-continuous Paths in a Poisson Target Field with I.I.D. Rewards}
\label{section:problem:problem2}

In this section, we present a two-dimensional special case. We believe this is the simplest case that captures all key aspects of the problem we presented above, namely: {\em (i)} the dynamics with drift, {\em (ii)} the stochastic nature of the environment, and {\em (iii)} the online (dynamic) nature of the task. This simple case is particularly relevant to the motivational example presented in Section~\ref{section:introduction} on information gathering. Furthermore, this case is also analytically tractable. 

{\bf Dynamics:} Let $X = \reals^2$, and define the dynamics governing the robot with the following ordinary differential equation:
\begin{align}
\begin{array}{c}
\dot{x}_1(t)  = v, \\
\dot{x}_2(t)  = u(t),
\end{array}
\label{eqn:dynamics}
\end{align}
where $[x_1(t) \; x_2(t)] \in \reals^2$ denotes the state of the robot, $v$ is a constant, and $|u(t)| \leq w$ is the control input. This robot travels with constant speed $v$ along the longitudinal direction ($x$-axis) and with bounded speed $u$ in the lateral direction ($y$-axis). We define the {\it agility} of the robot as $\agility = w  / v$. The larger this number $\alpha$, the more maneuverable the robot is. 

{\bf Targets and reward:} The target locations are generated by a two dimensional {\it Poisson point process} with intensity $\lambda$. That is, the number of targets $\Psi(A)$ for any region $A \in \reals^2$ follows a Poisson distribution, \ie, 
$
\Psi(A) \sim Poi(\lambda \, |A|).
$
The reward associated with each target is chosen from a common distribution independently. Let $r(p)$ denote the reward associated with the target at location $p \in \reals^2$. Then, $\{ r(p_i) : i \in \naturals\}$ are independent identically distributed random variables.

{\bf Target-sensing region:} The robot has a fixed target-sensing range $m$. That is, when the robot is at state $x(t) = [x_1(t) \; x_2(t)]$, it obtains the target's information, namely its location $p = [p_1  \;  p_2]$ and the associated reward $m$, for all targets located in 
$$
\mathcal{P}_\Psi \left(x(t)\right) = \left\{p \in \reals^2 \,:\,   0 \leq p_1 - x_1(t) \leq m \mbox{ and } (p, m) \in \Psi \right\}.
$$

%

\subsection{Data Gathering and Inference with Gaussian Noise}
\label{section:problem:data_gathering}

We established the maximum-reward motion in stochastic environments as a foundational problem. How is this problem related to data gathering with an agile robotic vehicle? 

In this section, we relate the maximum-reward motion problem with a special data-gathering problems as follows. 
We consider a robot tasked with estimating a fixed scalar $\theta$ from spatially-distributed measurements corrupted by Gaussian noise. As the robot navigates through the two-dimensional environment, it observes various candidate locations where measurements can be taken. The robot also observes the ``value of the information'' that these locations may potentially provide, if the robot visits that location and measures $\theta$. Based on the locations and their potential value of information, the robot must decide a subset of the locations to visit, and collect noisy measurements of the variable $\theta$ at each of those locations. 

The higher the value of information, the less noisy will be the observation at that location. 
Let the prior belief on $\theta$ be Gaussian-distributed with mean $\mu_0$ and variance $\frac 1 {\beta_0}$, \ie,
$$
\theta \sim \mathcal{N}(\mu_0, 1/{\beta_0}).
$$
%
Let the likelihood function of measurement $y_i$ of $\theta$ also be Gaussian distributed, centered at $\theta$ with variance $\frac 1 {\beta_i}$, \ie,
\begin{align}
y_i | \theta \sim \mathcal{N}(\theta, \frac 1 {\beta_i}). \label{eqn:Gaussian_measurement}
\end{align}
Notice that $\beta_i$ is the precision of measurement $y_i$. 
Given sensor measurements $\mathbf{y} = [y_1, y_2, \dots, y_n]$, we can derive the posterior probability of $\theta$ conditioning on $\mathbf{y}$ using Bayes' rule,
$$
\theta | \mathbf{y} \sim \mathcal{N}(\mu_n, 1/{\beta_n'}),
$$ 
with the updated mean $\mu_n$ and variance $\frac 1 {\beta_n'}$ satisfying
\begin{align}
\beta_n' = \beta_0 + \beta_1 + \dots + \beta_n.
\label{eqn:2-1}
\end{align}

Suppose the potential sensing locations are randomly distributed in the environment, and the robot is tasked with estimating $\theta$, subject to the differential constraints given by Equation~\eqref{eqn:dynamics}. 
The robot does not know the precise locations where measurements can be taken, but instead these target locations are discovered on the fly. Once a target location $p_i$ enters the target-detection region of the robot, the robot observes the precision $\beta_i$ of the corresponding measurement at $p_i$. If the robot chooses the visit $y_i$, then it will measure $\theta$, where the measurement is corrupted with Gaussian noise of variance $1/\beta_i$.

The robot is assigned the task of estimating $\theta$ as best as possible as its navigates through the field, and its performance is measured by the variance of the posterior distribution (the lower the better).
This problem is a specific instance of the maximum-reward motion problem defined in Section~\ref{section:problem:problem2}. In this setting, the ``reward'' associated with sensing location $y_i$ is precisely $\beta_i$. 
This problem is also motivated by the selection of unattended ground sensors, which we discuss as an application in Section~\ref{section:sensor-selection}.
However, in Sections~\ref{section:preliminaries} and \ref{section:analysis}, where we present our main results, we will focus on the maximum-reward problem, since we believe that the maximum-reward problem represents a more general setting.


%% file: preliminaries.tex

\section[Preliminaries: Last Passage Percolation and the Analysis of Motion on State Lattices]{\texorpdfstring{Preliminaries: Last Passage Percolation\\ And the Analysis of Motion on State Lattices}}
\label{section:preliminaries}

In this section, we develop some preliminaries. Specifically, we introduce a discrete problem that approximates the continuous problem of Section~\ref{section:problem}, and we devote this section to the analysis of this discrete problem. Many of our main results presented in the next section are obtained as the limiting cases of our results for the discrete problem in this section.

The approximate problem is constructed by using a lattice-based discretization. 
Note that lattice-based motion planning algorithms have long been widely adopted in robotics applications~\cite{Urmson:2008fw, Koenig:2004ut, pivtoraiko2011kinodynamic, cirillo2014lattice}. These algorithms form a directed lattice in the state space of the robot and select the optimal path through this lattice. This task is often computationally efficient, making it a practical approach even for challenging problem instances. 

We analyze this discrete problem and the resulting lattice-based planning algorithm, by establishing connections between this class of problems and a class of problems in non-equilibrium statistical mechanics. Roughly speaking, we view the robot as a particle traveling in a stochastic field. This perspective allows us to directly apply some of the recent results from the last-passage percolation problem~\cite{rolla2008last, zeng2013directed, Hambly:2007wm, martin2006last}. 

In what follows, we describe the lattice-based discretization in Section~\ref{section:preliminaries:problem3}. We introduce a lattice-based planning algorithm in Section~\ref{section:preliminaries:algorithm}. We analyze the fundamental limits of the problem in Section~\ref{section:preliminaries:analysis_unlimited_range} and the performance of the iterative planning algorithm in Section~\ref{section:preliminaries:analysis_limited_range}.

\subsection{The Last Passage Percolation Problem} \label{section:preliminaries:last_passage_percolation}
\begin{figure}[htbp]
        \begin{subfigure}[b]{0.2\textwidth}
        	        \centering
                \includegraphics[height=3.2cm]{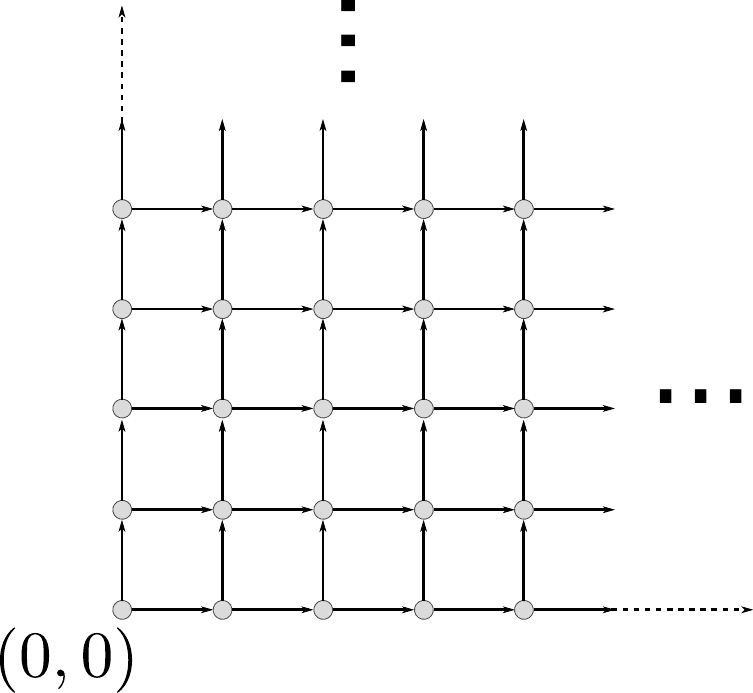}
                \caption{The two-dimensional lattice.}\label{figure:two_dim_lattice}
        \end{subfigure}\qquad
                \begin{subfigure}[b]{0.24\textwidth}
        	        \centering
                \includegraphics[height=1.6cm]{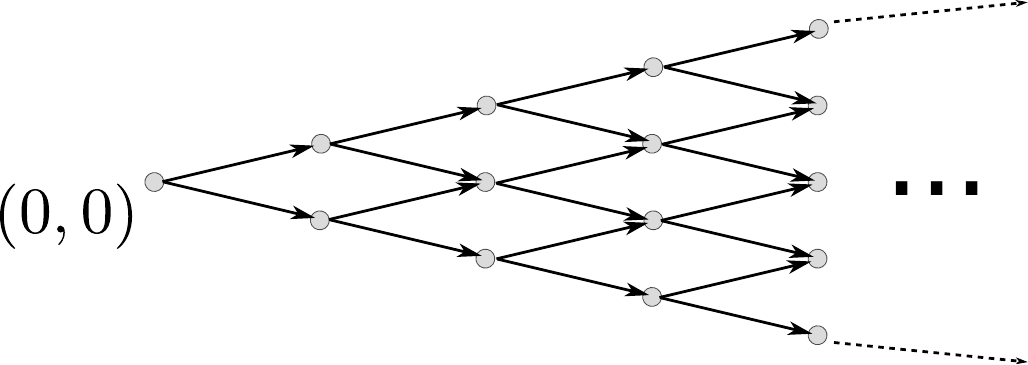}
                \vspace{0.13in}
                \caption{Two-dimensional lattice as the discretization of Equation~\eqref{eqn:dynamics}.} \label{figure:dynamic_lattice}
        \end{subfigure}
	\vspace{0.3in}
	
        \begin{subfigure}[b]{0.5\textwidth}
        	        \centering
                \includegraphics[height=2.2cm]{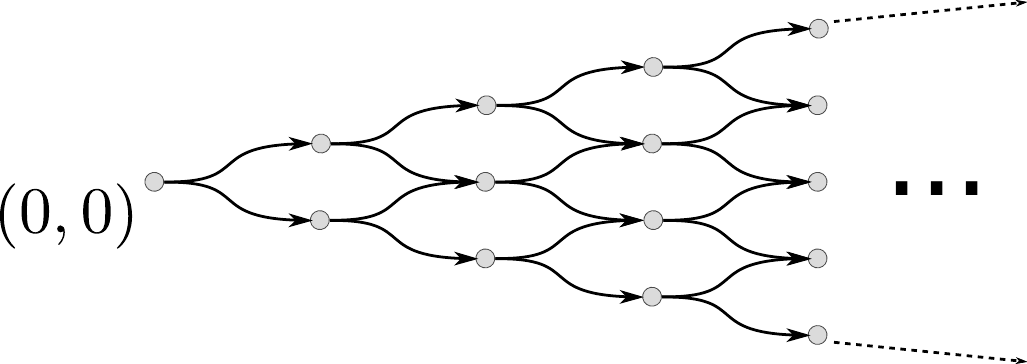}\hspace{0.35in}
                \caption{A lattice with Dubins paths.} \label{figure:dubins_lattice}
        \end{subfigure}
        \caption{The two-dimensional directed regular lattice, $\naturals^2$, is illustrated in Figure (a). An example state-lattice for a curvature-constrained Dubins vehicle, is shown in Figure (b). The latter lattice can be embedded in $\naturals^2$.} 
\end{figure}

A graph $(V,E)$ is called a {\em $d$-dimensional regular lattice}, if $V = \naturals^d$, and $(v,v') \in E$ if and only if $v = (v^1, v^2, \dots, v^d)$ and $v' = (v^1, v^2, \dots, v^{k-1}, v^k + 1,  v^{k+1}, \allowbreak \dots, v^d)$ for some $k$. The two-dimensional regular lattice is illustrated in Figure~\ref{figure:two_dim_lattice}.

Let $\vert v  \vert$ denote the distance of the vertex $v$ from the origin, measured by the number of vertices that any path between these two vertices has to visit. In other words, for the vertex $v$ as a $d$-dimensional vector of natural numbers, \ie, $v = (\vertexnumber{1}, \vertexnumber{2}, \dots, \vertexnumber{v}) \in \naturals^d$, we know that 
$
\vert v \vert :=  \vert \! \vert v \vert \! \vert_1 = \sum_{j=1}^d \vertexnumber{j}.
$

Let $\pathset(v_0, v)$ denote the set of all possible paths that start from vertex $v_0 \in V$ and end at vertex $v \in V$. With a slight abuse of notation, let $\pathset(v_0,n)$ denote the set of all paths that start at vertex $v_0$ and cross exactly $n$ vertices. When $v_0$ is the origin, \ie, $v_0 = \mathbf{0}$, we drop it from the notation, and we simply write $\pathset(v)$ for the set of all paths that start from the origin and end at vertex $v$. 
Then, the maximum total reward starting from $v_1$ and reaching $v$ on a $d$-dimensional regular lattice is defined as: 
$$
\optimaltotalreward (v_0,v) := \max_{\pi \in \pathset(v_0,v)} \,\,\sum_{v \in \pathname} r(v),
$$
where we write $v \in \pi$, when a path $\pi$ crosses a vertex $v$. 
We denote the maximum total reward by a path that starts from $v_1$ and crosses at most $n$ vertices as: 
$$
\optimaltotalreward (v_0,n) := \max_{\pi \in \pathset(v_0,n)} \,\,\sum_{v \in \pathname} r(v).	
$$
Finally, when $v_0 = \mathbf{0}$ (i.e., when the paths start at the origin), we simply drop it from the notation: We write $\optimaltotalreward(v)$ and $\optimaltotalreward(n)$ for $\optimaltotalreward(\mathbf{0}, v)$ and $\optimaltotalreward(\mathbf{0}, n)$, respectively. 

If a vertex $v$ is exactly $n$ steps away from the origin, \ie, $\vert v_\mathrm{dest} \vert = n$, then $\Pi(v_\mathrm{dest}) \subset \Pi(n)$. Since $\Pi(v_\mathrm{dest})$ is a subset, it is obvious that its maximum reward cannot exceed that of $\Pi(n)$. In other words, $\optimaltotalreward (v_\mathrm{dest}) \leq \optimaltotalreward (n)$.


\subsection{Problem 3: Maximum-reward Motion on State Lattices} \label{section:preliminaries:problem3}

In this section, we formulate a discrete problem that resembles Problem 2 of Section~\ref{section:problem:problem2}. This new problem indeed captures a discretized version of Problem 2. However, we stress that the problem presented in this section is {\em not} a special case. 

{\bf Dynamics:}  
%
%
A $d$-dimensional directed regular state lattice $L_d = (V,E)$ is a graph that satisfies: {\em (i)} $V$ is a countable set of vertices such that each vertex $v\in V$ is a state of the vehicle; {\em (ii)} $E \subset V \times V$ is a set edges, such that for all $(v,v')\in E$, there exists a dynamically-feasible trajectory $x_e : [0,\finaltime_e] \to X$ that connects $v$ and $v'$, \ie, $x(0) = v_1$ and $x(\finaltime_e) = v_2$; {\em (iii)} $L_d$ is isomorphic to a $d$-dimensional regular lattice. 

Therefore, the dynamics of Equation~\eqref{eqn:dynamics} can be discretized as a two-dimensional directed regular state-lattice, as given in Figure~\ref{figure:dynamic_lattice}. An example state-lattice for a non-holonomic vehicle is shown in Figure~\ref{figure:dubins_lattice}. Both examples are isomorphic to the two-dimensional regular lattice in Figure~\ref{figure:two_dim_lattice}. 

{\bf Targets and reward:} Each vertex $v \in V$ is associated with an independent, identically distributed random reward $r(v)$.

{\bf Target-detection range:} The target-detection range is a positive number $m$. Any vertex reachable with a path of length $m$ is considered to be within the target-detection range. That is, the vehicle can observe the reward $r(v)$ for all vertices that are within a distance of $m$ to the vehicle. 

{\bf Task:} The vehicle is again tasked with moving through its environment and maximizing the total reward collected. 

From Figure~\ref{figure:dynamic_lattice}, we see that Problem 3 of this section is a discrete version of the Problem 2 of Section~\ref{section:problem:problem2}. We stress that Problem 3 of this section can be more general. For example, the state lattice in \ref{figure:dubins_lattice} represents the Dubins vehicle dynamics. 

In the rest of Section~\ref{section:preliminaries}, we analyze Problem 3. This analysis will be used for deriving our main results for Problem 2. These results will be presented in Section~\ref{section:analysis}.

\subsection{Iterative Motion Planning Algorithm for State Lattices} \label{section:preliminaries:algorithm}
A path on $G = (V,E)$ is a sequence of vertices, $(v_1, v_2,\allowbreak  \dots, v_k)$, such that consecutive vertices are connected with an edge, \ie, $(v_i, v_{i+1}) \in E$ for all $i \in \{1, 2, \dots, k-1\}$. The set of all paths on $G$ that starts at vertex $v$ is denoted by ${\tt Paths}(v)$.

We consider the following motion planning algorithm. 
Suppose the vehicle starts at an initial state $z_\mathrm{init} \in V$. In each iteration, the maximum-reward path, say $(v_1,v_2,\dots, v_k)$, within the ``visible'' portion of the lattice is computed, and the vehicle executes the resulting dynamically-feasible trajectory until its end. The same procedure is repeated, after the vehicle reaches the final state $v' = x_{e_k}(\finaltime_{e_k})$. 

We call this algorithm the {\em iterative lattice-based online motion planning algorithm}, which we formalize in Algorithm~\ref{algorithm:lattice}. The ${\tt GetCurrentState}()$ procedure returns the current state of the robot, and ${\tt Execute} (x)$ refers to the command that makes the robot follow the trajectory $x$. The algorithm first retrieves the robot's current state (Line~\ref{line:sense}). Subsequently, it observes the reward associated with the vertices for the visible region of the lattice (Line~\ref{line:project}). It then searches for the optimal path over this region (Line~\ref{line:optimize}). Finally, it executes this path until the vehicle reaches the end of the path (Line~\ref{line:actuate}). This procedure continues for $N$ iterations (Lines~\ref{line:loop_begin}-\ref{line:loop_end}).

\begin{algorithm}[htbp]
\begin{algorithmic}[1]\small
\For{$t = 1, \dots, N$} \label{line:loop_begin} 
	\State $\text{state} \leftarrow {\tt GetCurrentState}()$\; \label{line:sense}
	\State ${\tt PerceiveEnvironment}()$\; \label{line:project}
	\State $\pi \gets \arg\max_{\pi}\{ \totalreward(\pi) \!:\! \pi \in {\tt Paths}(\text{state})\}$\label{line:optimize}
	\State ${\tt Execute}(\pi)$\; \label{line:actuate}
\EndFor\label{line:loop_end}
\caption{Iterative lattice-based online motion planning}\label{algorithm:lattice}
\end{algorithmic}
\end{algorithm}

In Line~\ref{line:optimize}, the algorithm computes the maximum-weight path on a finite weighted graph. Let us note that this problem is NP-hard in general~\cite{Schrijver:2003tv}. However, the problem can be solved efficiently on acyclic graphs~\cite{Schrijver:2003tv}.\footnote{Acyclic graphs arise in lattice-based motion planning, for instance, when the robot does not return to previously visited locations, \ie, the robot constantly explores new regions in the environment. When lattice-based motion planning algorithms are applied to robots subject to substantial drift, the resulting lattice is also often acyclic. It is easy to show that any state lattice for the system presented in Section~\ref{section:problem:problem2} is acyclic.}

In what follows, we analyze the performance of Algorithm~\ref{algorithm:lattice} on the $2$-dimensional directed regular state-lattice (and more generally, $d$-dimensional regular lattices). 







\subsection{Mean Reward with Unlimited Sensing Range}
\label{section:preliminaries:analysis_unlimited_range}

In this section, we analyze the maximum reward that robot can collect, if it had infinite sensing range. 
The results of this section can be regarded as fundamental limits: Even when the robot has infinite sensing range, the reward it can collect is bounded by what we report in this section. 

For our analysis, we focus on the maximum reward collected per unit distance traveled, which we call the {\em mean reward}.
We define the {\em maximum mean reward} as follows:
$$
\optimalmeanreward(n) := \frac{\optimaltotalreward(n)}{n}. 
$$
First, we show that the limit $\expectedoptimalmeanreward_d:= \lim_{n\to\infty}\EE[\optimalmeanreward(n)]$, which we call {\em expected maximum mean reward}, is well defined. 
\begin{theorem} 
\label{theorem:lattice-limit-exists}
The following holds:
$$
\lim_{n\to\infty} \EE[\optimalmeanreward(n)] = \lim_{n\to\infty} \frac{\EE[\optimaltotalreward(n)]}{n} = \sup_{n \in \naturals} \frac{\EE[\optimaltotalreward(n)]}{n} \in \reals \cup \{ \infty \} .
$$
\end{theorem}

The proof can be found in Appendix~\ref{app:lattice-limit-exists}. For general distributions, the value of $\expectedoptimalmeanreward_d$ can not be computed directly. However, we can compute asymptotics for special cases of light-tailed (exponential and geometric distributions, specifically) and heavy-tailed distributions. 

A distribution is light tailed if its tail is bounded by an exponentially decreasing function. For example, Gaussian, geometric, exponential and all bounded distributions are light-tailed, while Pareto, logarithmic normal, Cauchy and Student's t distributions are heavy-tailed. More precisely: 
\begin{definition} [See \cite{rolski2009stochastic}] 
\label{definition:light_tails}
The distribution $F$ is said to be light-tailed, if it has an exponentially bounded tail, \ie, for some $a, b>0$, we have
$
1 - F(x) \leq a e^{- bx}, \text{ for all } x > 0.
$
A heavy-tailed distribution is one that is not light-tailed.
\end{definition}

\subsubsection{Light-tailed reward distributions}
The following theorem shows that, when the reward distribution is a light-tailed distribution, the maximum mean reward is bounded.
\begin{theorem} [See Theorem 4.1 in~\cite{martin2006last}]
\label{theorem:lattice-optimal-light-tailed}
Suppose the rewards $r$ are independent and identically distributed with distribution $F$, such that 
\begin{equation} 
\int_0^\infty  (1 - F(s))^{1/d} ds < \infty,
\label{eqn:light-tail-reward-definition}
\end{equation} 
then $\expectedoptimalmeanreward_d$ is finite for any $d \ge 2$, \ie,
$$
\expectedoptimalmeanreward_d = \lim_{n\to\infty}\EE[\optimalmeanreward(n)] = \lim_{n \to \infty} \EE \left[ \frac{T^*(n)}{n} \right ] = \nextconstant
$$
for some finite constant $\thisconstant \in \reals$.
\end{theorem}
The proof can be found in Appendix~\ref{app:lattice-optimal-light-tailed}. Note that Equation~\ref{eqn:light-tail-reward-definition} is only very slightly stronger than the existence of a finite $d^\text{th}$ moment, \ie, $\EE[r^d] < \infty$. Moreover, when the dimension is $d=2$ and if rewards follow exponential or geometric distributions (both satisfy Equation~\ref{eqn:light-tail-reward-definition}), then $\expectedoptimalmeanreward_2$ can be computed explicitly.

\begin{proposition}
\label{proposition:lattice-exponential-geometric}
For exponential and geometric reward distributions with mean $\mu$ and variance $\sigma^2$ on a two-dimensional lattice, the expected maximum mean reward $\expectedoptimalmeanreward_2$ can be computed explicitly
\begin{align*}
\expectedoptimalmeanreward_2 = \mu + \sigma.
\end{align*}
\end{proposition}
The proof is given in Appendix~\ref{app:lattice-exponential-geometric}.

\subsubsection{Heavy-tailed reward distributions}
We just showed that, for all light tailed distributions, $\expectedoptimalmeanreward_2$ is finite. In contrast, the following theorem states that, when the reward distribution is heavy-tailed, $\expectedoptimalmeanreward_2$ is infinite.
\begin{theorem} [See Proposition 2 in~\cite{Angel2012}]
\label{theorem:lattice-optimal-heavy-tailed}
Suppose the rewards are independent and identically distributed and $\EE [r^d] = \infty$, then $\expectedoptimalmeanreward_d = \infty$.
\end{theorem}
Here we consider a specific instance of the heavy-tailed distribution family, the \emph{Pareto distribution}, which is commonly used to describe the allocation of wealth among individuals or distribution of income. More specifically, 
\begin{definition} [Pareto Distribution]
\label{def:pareto}
The Pareto distribution with index parameters $x_m$ and $\alpha$ is defined as follows:
$$
\PP(X \le x) = 
\begin{cases}
1 - \left(\frac{x_m}{x}\right)^{\alpha}, & \ x \ge x_m \\
0, & \ x < x_m
\end{cases}
$$
\end{definition}
For a Pareto distribution, more accurate results regarding the growth rate of $\optimaltotalrewardtwo(n)$ can be obtained as follows:
\begin{proposition}
\label{proposition:lattice-optimal-pareto}
Suppose the rewards $r$ are independent and Pareto distributed with parameter $\alpha \in (0,2)$. Then, the optimal mean reward $\expectedoptimalmeanreward_2$ is infinite. 
Moreover, the growth rate of $\optimaltotalrewardtwo(n)$ is at the order of $n^{(2/\alpha)}$, \ie,
\begin{equation*}
\begin{split}
\optimaltotalrewardtwo(n) = O \left( n^{2/\alpha} \right), \\
\optimalmeanrewardtwo(n) = O \left( n^{2/\alpha - 1} \right).
\end{split}
\end{equation*}
\end{proposition}
The proof is given in Appendix~\ref{app:lattice-optimal-pareto}.
\subsection{Mean Reward with Limited Sensing Range}
\label{section:preliminaries:analysis_limited_range}

In this section, we consider robots with a limited sensing range. Suppose the sensing range is $m$. To travel a distance of $n$ vertices, we follow the best path for $m$ steps, and then repeat this procedure, until the $n$th vertex is reached. 
We are particularly interested in comparing the reward collected with limited sensing range in this way to the reward collected with unlimited sensing range as described in the previous section. 

Let $\totalreward_1$ denote the maximum total reward collected by a path that starts from the origin vertex, $\mathbf{0}$, and has length $m$, \ie, $\totalreward_1 := \optimaltotalreward(\mathbf{0}, m)$. Let $v_1$ denote the vertex where the maximum-reward path achieving reward $\totalreward_1$ ends. More generally, define $\totalreward_k := \optimaltotalreward(v_{k-1}, m)$, where $v_{k}$ is the vertex the path achieving reward $\totalreward_{k-1}$ ends. Assume $n$ is a multiple of $m$. Define:
\begin{equation}
\label{eqn:definition-T-IMP}
\IMPtotalreward(n;m) := \sum_{i = 1}^{n/m} \totalreward_i.
\end{equation}

Finally, define the mean reward collected by the robot in $n$ steps with sensing range $m$ as:
$$
\IMPmeanreward(n;m) = \frac{\IMPtotalreward(n;m)}{n}.
$$

In our analysis, we compare mean reward $\IMPmeanreward(n;m)$ with $\expectedoptimalmeanreward_d$. Recall that the former is the mean reward that the robot can collect with limited sensing range $m$, and the latter is the mean reward with unlimited sensing distance. 


Surprisingly, the difference in performance between the unlimited versus limited sensing range turns out to be drastically different, when the distribution of the reward is light tailed versus heavy tailed. We analyze both cases in this order.

\subsubsection{Light-tailed reward distributions}
Theorem~\ref{theorem:lattice-sensing-light-tailed} shows that when the reward is in the light-tailed family, iterative motion planning algorithms achieve near-optimal performance even with very limited sensing range.

\begin{theorem}
\label{theorem:lattice-sensing-light-tailed}
Suppose the rewards $r$ are independent, identically distributed and satisfy Equation~\ref{eqn:light-tail-reward-definition}.
Then, for any $\delta > 0$, there exists a constant $c$ such that $\IMPmeanreward(n,c\log n)$ converges to $\expectedoptimalmeanreward_d$ in probability, \ie, 
$$
\lim_{n \to \infty}\PP\left( \,\big\vert \IMPmeanreward(n,c\log n)  - \expectedoptimalmeanreward_d \big\vert \,\ge\, \delta \,\right) \,\,=\,\, 0.
$$
\end{theorem}

The proof of Theorem~\ref{theorem:lattice-sensing-light-tailed} is provided in Appendix~\ref{app:lattice-sensing-light-tailed}. Roughly speaking, Theorem~\ref{theorem:lattice-sensing-light-tailed} implies that the robot can navigate to any vertex that is $n$ steps away almost optimally (as if it had infinite sensing distance), even when its sensing range is only $c\,\log n$. 
This result is remarkable, as $\log n$ is much smaller than $n$. Our simulation results provided in Section~\ref{sec:experiments} support our conjecture when $d=2$.

\subsubsection{Heavy-tailed reward distributions}
\label{section:preliminaries-heavy}

We consider the case where the rewards follow a Pareto distribution on a two-dimensional regular lattice, \ie, $d=2$. We show that the iterative motion planning algorithm can {\em not} achieve a near-optimal performance with limited sensing distance $o(n)$. 

\begin{theorem}
\label{theorem:lattice-sensing-pareto}
Suppose the assumptions of Proposition~\ref{proposition:lattice-optimal-pareto} hold. 
%
Then, there exists a probability space $(\Omega, \mathcal{F}, P)$ such that when $M(n)$ is a sub-linear function of $n$, \ie, $\lim_{n \to \infty} M(n)/n = 0$, we have
$$
\lim_{n \to \infty}\frac {\IMPmeanreward \left( n; M(n) \right)} {\optimalmeanrewardtwo(n)} = 0.
$$
\end{theorem}
See Appendix~\ref{app:theorem-heavy-discrete} for the proof. 
Roughly speaking, Theorem~\ref{theorem:lattice-sensing-pareto} states that, if the sensing range of the robot is slightly less than $n$, then the reward collected will be much lower.  

The findings of this section are summarized below. 
\begin{remark}
According to Theorem~\ref{theorem:lattice-sensing-light-tailed}, when the reward follows a light-tailed distribution, $c\, \log n$ sensing range is adequate to navigate optimally (as if the robot had infinite sensing range). However, according to Theorem~\ref{theorem:lattice-sensing-pareto}, when the reward distribution follows the Pareto law (a heavy-tailed distribution), any non-negligible limitation in sensing range leads to substantial losses in performance. 
\end{remark}

%% file: analysis-continuous.tex

\section{Analysis of the Continuous Problem}
\label{section:analysis}
In this section, we return to Problem 2 (see Section~\ref{section:problem:problem2}), namely the maximum-reward motion problem in $\reals^2$. We study this problem, assuming (i) unit agility, (ii) infinite sensing range, and (iii) infinite computation capability. We find the necessary requirements on agility, sensing range, and computation capabilities that will allow the robot to perform close to these fundamental limits

Our analysis is based on our results presented in Section~\ref{section:preliminaries}. Specifically, we show that the continuous problem is the limiting case of the discrete Problem 3 (see Section~\ref{section:preliminaries:problem3}).

This section is organized as follows. In Section~\ref{sec:continuous-fundamental-limits} we analyze the fundamental limits of robot performance, given infinite sensing range with unit agility. We introduce the iterative motion planning algorithm in Section~\ref{sec:continuous-algorithm} and study its performance under limited sensing range in Section~\ref{sec:continuous-sensing} for different types of reward distributions. In Section~\ref{sec:agility} we move on to requirements on robot agility. In Section~\ref{section:computational_workload} we study the computational workload for motion planning and inference tasks, respectively.

\subsection{Fundamental Limits: The Analysis of Mean Reward with Infinite Sensing Range and Unit Agility}
\label{sec:continuous-fundamental-limits}

Let $\Pi(L)$ be the set of all feasible paths that start from the origin and travels a distance of $L$ in the longitudinal direction, \ie, the $x_1$ axis. 
Recall the assumption that the reward locations $\{p_i\}$ are generated by a Poisson point process with intensity $\lambda$. The amount of reward at each target is an i.i.d. random variable $r (p_i)$ that follows some common reward distribution. 

Let $\optimaltotalreward(L)$ denote the optimal total reward collected by following any path in $\Pi(L)$ with infinite sensing distance, \ie,
$$
\optimaltotalreward (L) := \max_{\pi \in \Pi(L)} \,\,\sum_{p_i \in \pi} r(p_i).
$$
Let $\optimalmeanreward(L)$ denote the optimal mean reward collected in the same manner, \ie, 
$$
\optimalmeanreward (L) := \frac{\optimaltotalreward(L)}{L}.
$$

Throughout this section, we assume unit robot agility $\agility = 1$, and we analyze the total reward that the robot can collect.

The first two results for the continuous problem are extensions of Theorem~\ref{theorem:lattice-limit-exists}-\ref{theorem:lattice-optimal-light-tailed}.

\begin{theorem}[Well-posedness of the Mean Reward]
\label{theorem:cont-limit-exists}
Suppose the reward locations are generated by a Poisson point process with intensity $\lambda$ on $\reals^2$. The reward associated with each target is chosen from a common distribution $F$ independently. The robot dynamics satisfies the following ordinary differential equation:
$$
\dot{x}_1(t)  = v, \quad \dot{x}_2(t)  = u(t), \quad \vert u(t) \vert \leq v.
$$
Then,
$$
\lim_{L \rightarrow \infty} \frac {\EE \optimaltotalreward (L)} {L} = \sup_L \frac {\EE \optimaltotalreward(L)} L.
$$
\end{theorem}

The proof for both Theorem~\ref{theorem:cont-limit-exists} is given in Appendix~\ref{app:cont-limit-exists}. For simplicity of notation we will define this optimal mean reward as
$$
\expectedoptimalmeanreward_2 := \sup_L \frac {\EE \optimaltotalreward(L)} L.
$$

Next, we compute asymptotics for the optimal mean reward. As in the discrete case, we consider the light-tailed and heavy-tailed reward distributions separately. 

\subsubsection{Light-tailed reward distributions}
The following theorem is the continuous counterpart of Theorem~\ref{theorem:lattice-optimal-light-tailed}, and it shows that, when the reward distribution is light-tailed, the maximum mean reward is bounded.
\begin{theorem}[Mean Reward Asympototics for Light-tailed Rewards]
\label{theorem:cont-optimal-light-tailed}
Suppose the conditions of Theorem~\ref{theorem:cont-limit-exists} hold, and the rewards $r(v)$ are independent, identically distributed, and satisfy Equation~\ref{eqn:light-tail-reward-definition}. Then, $\expectedoptimalmeanreward_d$ is finite for any $d \ge 2$, \ie,
$$
\expectedoptimalmeanreward = \lim_{n\to\infty}\EE[\optimalmeanreward(n)] = \lim_{n \to \infty} \frac{\EE[T^*(n)]}{n} = \nextconstant
$$
for some finite constant $\thisconstant \in \reals$.
\end{theorem}
Interested reader please refer to the proof for Theorem 1.2.1 in~\cite{cator2011hydrodynamical}, which applies the subadditive ergodic theorem. 

\subsubsection{Heavy-tailed reward distributions}
In contrast, the following theorem shows that, when the reward distribution is heavy-tailed, $\expectedoptimalmeanreward_2$ is infinite. This theorem is the continuous counterpart of Theorem~\ref{theorem:lattice-optimal-heavy-tailed}.

\begin{theorem}[Mean Reward for Heavy-Tailed Rewards]
\label{theorem:cont-optimal-heavy-tailed}
Suppose the conditions of Theorem~\ref{theorem:cont-limit-exists} hold and the rewards satisfy $\EE [r^2] = \infty$. Then, the optimal mean reward $\expectedoptimalmeanreward_2$ is infinite. 
\end{theorem}

The proof is in Appendix~\ref{app:cont-optimal-heavy-tailed}. Similar to the case with discrete lattices, more accurate results can be derived for the Pareto distributions.
\begin{proposition}[Mean Reward Asympototics for Pareto Rewards]
\label{proposition:cont-optimal-pareto}
Suppose the conditions of Theorem~\ref{theorem:cont-limit-exists} hold, and the rewards $r$ are Pareto-distributed with parameter $\alpha \in (0,2)$. Then, the optimal mean reward $\expectedoptimalmeanreward$ is infinite. Moreover, the growth rate of $\optimaltotalrewardtwo(L)$ is order $L^{(2/\alpha)}$, \ie, 
\begin{equation*}
\begin{split}
\optimaltotalrewardtwo(L) = O \left( L^{2/\alpha} \right), \\
\optimalmeanrewardtwo(L) = O \left( L^{2/\alpha - 1} \right).
\end{split}
\end{equation*}
\end{proposition}
The proof is given in Appendix~\ref{app:cont-optimal-pareto}.

\subsection{The Iterative Motion Planning Algorithm}
\label{sec:continuous-algorithm}
Similar to the planning algorithm on discrete lattices, in the continuous space the planning algorithm proceeds in an iterative manner. Suppose the robot starts at an initial state $z_\mathrm{init}$. First, the best feasible trajectory $x_{e} : [0,T_{e}] \to X$ within the ``visible'' region of the lattice is computed, and the robot follows this dynamically-feasible trajectory until the end. After the robot completes this trajectory, the same procedure is repeated. This algorithm is formalized in Algorithm~\ref{algorithm:continuous}.

Let ${\tt PerceiveEnvironment}()$ (Line~\ref{line:perceive}) be a procedure that returns the set of targets/rewards that are visible to the robot. The robot then computes the optimal path within the set of trajectories ${\tt Paths}(\text{state})$ to maximize the total reward collected (Line~\ref{line:compute}). In this problem, ${\tt Paths} = \{ \pi: \dot{x_1}=v, |\dot{x_2}| \leq w\}$. The procedure ${\tt Execute} (\pi)$ (Line~\ref{line:execute}) commands the robot to move along the planned path $\pi:[0, m/v] \to X$ and returns the total reward collected along this path. 
After completion of this command, the entire procedure is repeated until time $distance_x$ is greater than travel distance $L$ (Lines~\ref{line:while_begin}-\ref{line:while_end}). 

\begin{algorithm}[htbp]
\begin{algorithmic}[1]\small
\State ${\tt distance}_x \gets 0$
\While{${\tt distance}_x < L$} \label{line:while_begin}
   \State $\text{state} \leftarrow {\tt GetCurrentState}()$\;
	\State ${\tt PerceiveEnvironment}()$\;  \label{line:perceive}
	\State $\pi \gets \arg\max\{ \totalreward(\pi):\pi \in {\tt Paths}(\text{state}) \}$\; \label{line:compute}
	\State $\totalreward_i \gets {\tt Execute}(\pi)$\; \label{line:execute}
	\State $Q \gets Q + \totalreward_i$ \; 
	\State ${\tt distance}_x \gets {\tt distance}_x + m$\; 
\EndWhile \label{line:while_end}
\caption{Receding-horizon online motion planning} \label{algorithm:continuous}
\end{algorithmic}
\end{algorithm}

\subsection{Sensing Requirements}
\label{sec:continuous-sensing}
We denote the sensing distance by $\sensingrange$. Define $\totalreward_i$ is the amount of total reward collected during the $i^\text{th}$ iteration of Algorithm~\ref{algorithm:continuous}, and let $\IMPtotalreward(L; m)$ denote the total reward collected with Algorithm~\ref{algorithm:continuous} throughout the entire mission, \ie,  
$$
\IMPtotalreward(L;\sensingrange) := \sum_{i = 1}^{L/\sensingrange} \totalreward_i.
$$
Define the mean reward collected by the Algorithm~\ref{algorithm:continuous} by
$$
\IMPmeanreward(L;\sensingrange) := \frac{\IMPtotalreward(L;\sensingrange)}{L}.
$$

In this section, we analyze the mean reward collected by the algorithm for two different reward distributions: (i) when the rewards are almost-surely bounded, (ii) when the rewards follow the Pareto distribution. The former is a light-tailed distribution, whereas the latter is a heavy-tailed distribution. 

\subsubsection{Light-tailed reward distributions}
\label{sec:cont-sensing-light-tailed}
The following result extends Theorem~\ref{theorem:lattice-sensing-light-tailed} and shows that the receding horizon algorithm still has near-optimal performance even in the continuous problem, when the sensing distance $m$ is at the order of $\log L$.
\begin{theorem}[Sensing range requirements for light-tailed rewards]
\label{theorem:cont-sensing-light-tailed}
Suppose the conditions of Theorem~\ref{theorem:cont-limit-exists} hold. Then, for any $\delta > 0$, there exists some constant $\nextconstant > 0$ such that
$$
\lim_{m \to \infty}\PP\left( \,\left\vert \IMPmeanreward(L;\thisconstant \log L)  -\expectedoptimalmeanreward_2 \right\vert \,\ge\, \delta \,\right) \,\,=\,\, 0.
$$
\end{theorem}

\subsubsection{Heavy-tailed reward distributions}
The following theorem shows that, when the rewards follow the Pareto distribution (a heavy-tailed distribution), then the 

\begin{theorem}[Sensing range requirements for Pareto rewards]
\label{theorem:cont-sensing-pareto}
Suppose the assumptions of Proposition~\ref{proposition:cont-optimal-pareto} hold. 
Then, there exists a probability space $(\Omega, \mathcal{F}, P)$ such that, for any sub-linear function $\sensingrange(L)$, \ie, $\lim_{L\to\infty} \sensingrange(L)/L = 0$, we have
$$
\lim_{L\to \infty}\frac{\EE[\IMPmeanreward(L;\sensingrange(L))] - \EE[\optimalmeanreward(L)]}{L^{(2/\alpha)-1}} = \nextconstant
$$
for some positive constant $\thisconstant$.
\end{theorem}
To avoid repetition we don't provide a detailed proof for Theorem~\ref{theorem:cont-sensing-light-tailed}-\ref{theorem:cont-sensing-pareto}, since the proof techniques are identical to Theorem~\ref{theorem:lattice-sensing-light-tailed}-\ref{theorem:lattice-sensing-pareto}.

\subsection{Agility Requirements}
\label{sec:agility}
In this section, we examine how agility impacts the performance of the robot, measured by the total reward collected. The main result of this section is the following:


\begin{theorem}[Agility Requirements]\label{theorem:agility}
Suppose the reward locations are generated by a Poisson point process with intensity $\lambda$ on $\reals^d$. The robot dynamics satisfies the following ordinary differential equation:
$$
\dot{x}_1(t)  = v, \quad \dot{x}_2(t)  = u(t),
$$
where $\vert u(t) \vert \leq w$. Then for any finite $L > 0$, there exists a constant $\nextconstant>0$ that is independent of $A$ (but depends on $L$) such that 
$$
\E[\optimalmeanreward (L)] =  \thisconstant\sqrt{\agility},
$$
where $\agility = w/v$ is the agility of the robot.
\end{theorem}


\subsection{Computation Requirements} 
\label{section:computational_workload}
In this section, we analyze the computational capabilities required on board the robot to perform the inference and planning tasks for data gathering. 
Specifically, we first analyze the amount of computational operations required to run the planning algorithm presented in Algorithm~\ref{algorithm:continuous}. We then analyze the amount of computational operations required to run inference algorithms, \eg, in the setting described in Section~\ref{section:problem:data_gathering}.

\paragraph{On motion planning} 
The planning algorithm, presented in Algorithm~\ref{algorithm:continuous}, is called periodically. Recall that $\sensingrange$ is the sensing range of the robot and $L$ is the length of the mission. Then, the planning algorithm is called exactly $L/\sensingrange$ times over the course of the mission. 
The planning procedure itself is a dynamic programming algorithm that computes the optimal path on an acyclic graph with $N$ nodes, where $N$ is the number of targets that are within the sensing range of the robot when the algorithm is called. The dynamic programming requires $O(N^2)$ steps. 
The expected value of the number can be computed as follows. 
The area of the target-detection region is of order $\agility \sensingrange^2$. Since the target locations are Poisson distributed with intensity $\lambda$, we have 
$
\E[N] = O(\lambda \agility\sensingrange^2)
$
This yields $O\left((\lambda\agility\sensingrange^2)^2\right)$ computational operations each time the planning algorithm is called. 
We normalize this number by the time it takes the robot to traverse this distance, \ie, by $\sensingrange/v$.
Then, the asymptotic running time complexity for motion planning is
$$
\planningtime = O\left( \frac{(\lambda\agility \sensingrange^2)^2}{\sensingrange/v} \right) 
= O\left(\lambda^2 \agility^2 \sensingrange^3 v \right).
$$

\paragraph{On inference task} 
The robot must perform some form of inference each time it visits a target location, processing the data collected at the same location.
Hence, the computational complexity of inference tasks is the number of tasks visited. The number of targets visited is analyzed similarly to the amount of reward collected. Following the proof of Theorems\ref{theorem:cont-optimal-light-tailed} and~\ref{theorem:agility}, we find that the number of targets visited while traversing a distance $\sensingrange$ is $O(\sqrt{\agility} \, \sensingrange)$. We assume that there is a constant number of operations performed at each location for inference. Then, the total number sensing operations is $O(\sqrt{\agility}\,\sensingrange)$. We normalize this number with the time it takes to travel distance $\sensingrange$ to arrive at the number of computational operations per unit time devoted to inference. The time is $\sensingrange/v$. Hence, the computational runtime complexity of inference is 
$$
\interencetime = O\left(\frac{\sqrt{\agility}\,\sensingrange}{\sensingrange/v}\right) = O\left(\sqrt{\agility}\,v\right).
$$

%% file: experiments.tex
\section{Computational Experiments}
\label{sec:experiments}
This section is devoted to the results of Monte-Carlo simulation studies that verify our theoretical results in Sections~\ref{section:preliminaries} and \ref{section:analysis}. We consider the discrete and continuous problems separately, and in each case we study the robot performance with different reward distributions, including geometric, exponential, Bernoulli, and Pareto.

\subsection{Optimal Mean Reward on Discrete Lattices}

\begin{figure}[hbt]
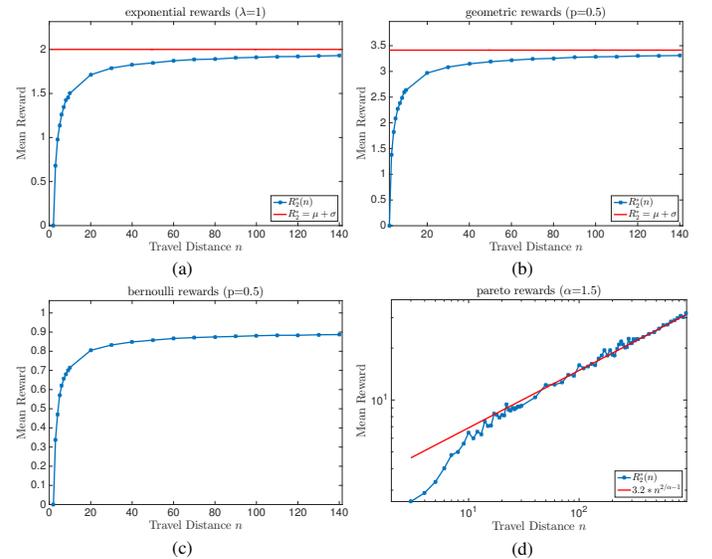

\myIncludeFourFigures
{data/reward_vs_length_lattice/{reward_vs_length_lattice.exponential.p=1.iter=1000}.eps}
{data/reward_vs_length_lattice/{reward_vs_length_lattice.geometric.p=0.5.iter=1000}.eps}
{data/reward_vs_length_lattice/{reward_vs_length_lattice.bernoulli.p=0.5.iter=1000}.eps}
{data/reward_vs_length_lattice/{reward_vs_length_lattice.pareto.p=1.5.iter=5000}.eps}
\caption{Optimal mean reward vs. travel distance on a two-dimensional regular lattice. The rewards associated with each vertex on the lattice are independent and identically distributed according to (a) exponential, (b) geometric, (c) Bernoulli and (d) Pareto distributions, respectively. (a)-(c) The mean reward for light-tailed distributions converges to a finite constant. (d) The mean reward for a heavy-tailed distribution goes to infinity.
}
\label{fig:lattice-optimal-mean-reward}
\end{figure}

In this section, we verify Theorems~\ref{theorem:lattice-limit-exists}-\ref{theorem:lattice-optimal-heavy-tailed} and Propositions~\ref{proposition:lattice-exponential-geometric}-\ref{proposition:lattice-optimal-pareto} in Monte-Carlo simulations. We consider a robot moving on a two-dimensional regular lattice, where the rewards are independent and identically distributed. We examine the optimal mean reward collected over the course of motion, and the results are shown in Figure~\ref{fig:lattice-optimal-mean-reward}. In this set of experiments, the rewards follow geometric, exponential, Bernoulli, and Pareto distribution, respectively. 

The exponential, geometric, Bernoulli distributions, shown in Figure~\ref{fig:lattice-optimal-mean-reward}(a)-(c), all belong the light-tailed family. Therefore, as predicted by Theorems~\ref{theorem:lattice-limit-exists}-\ref{theorem:lattice-optimal-heavy-tailed}, their mean reward converges quickly towards a finite constant. In addition, the mean reward of both the geometric and exponential distributions converges to $\expectedoptimalmeanreward_2 =\mu + \sigma$ (indicated by the red lines), which is the optimal mean reward predicted by Propositions~\ref{proposition:lattice-exponential-geometric}.

On the other hand, Figure~\ref{fig:lattice-optimal-mean-reward}(d) shows the log-log plot with rewards being Pareto-distributed (with parameter $\alpha=1.5$). The mean reward increased to infinitey with the travel distance, as dictated by Theorem~\ref{theorem:lattice-optimal-heavy-tailed} for all heavy-tailed distributions. Since Figure~\ref{fig:lattice-optimal-mean-reward}(d) is a log-log plot, the mean reward grows with travel distance $n$ at a rate of $n^{2/\alpha -1}$, as predicted by Proposition~\ref{proposition:lattice-optimal-pareto}. 

\subsection{Sensing Range on Discrete Lattices}
\label{sec:lattice-sensing}
\begin{figure}[hbt]
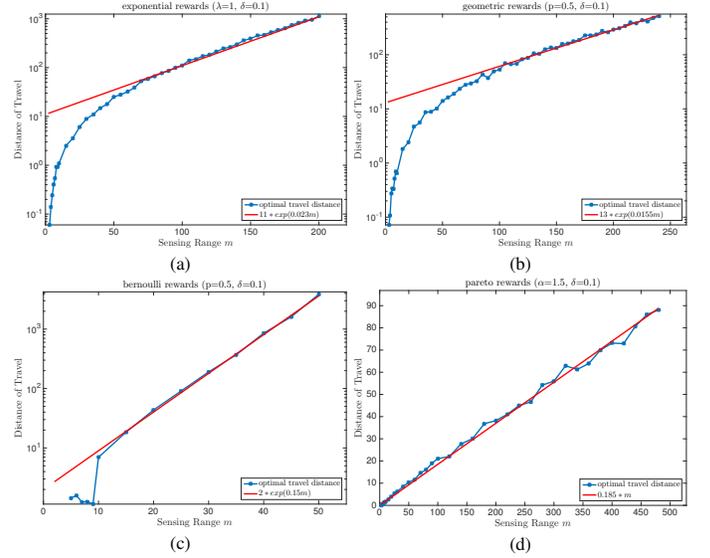

\myIncludeFourFigures
{data/dist_vs_horizon_lattice/{dist_vs_horizon_lattice.exponential.p=1.delta=0.1.iter=1000}.eps}
{data/dist_vs_horizon_lattice/{dist_vs_horizon_lattice.geometric.p=0.5.delta=0.1.iter=1000}.eps}
{data/dist_vs_horizon_lattice/{dist_vs_horizon_lattice.bernoulli.p=0.5.delta=0.1.iter=1000}.eps}
{data/dist_vs_horizon_lattice/{dist_vs_horizon_lattice.pareto.p=1.5.delta=0.1.iter=5000}.eps}
\caption{Average distance-of-travel versus sensing range on a two-dimensional regular lattice.
 (a)-(c) are log-linear plots for exponential, geometric and bernoulli rewards, respectively. The distance-of-travel grows exponentially fast with sensing range when reward distribution is in the light-tailed family. (d) is a linear plot for pareto rewards (in the heavy-tailed family), and the distance-of-travel only grows linearly with sensing range.
}
\label{fig:lattice-sensing}
\end{figure}

In this section, we verify Theorem~\ref{theorem:lattice-sensing-light-tailed}-\ref{theorem:lattice-sensing-pareto} in Monte-Carlo simulations. 
We consider a robot moving on a state lattice with limited sensing range $m$. It runs Algorithm~\ref{algorithm:lattice}. 
We fix the error level $\delta=0.1$. We let the robot travel with Algorithm~\ref{algorithm:lattice} until the mean reward it collects during one iteration of the receding-horizon algorithm is at least $\delta$ less than the optimal. In other words, the robot stops if $\totalreward_i/m < \optimalmeanrewardtwo - \delta$ for the $i^\text{th}$ iteration. In Figure~\ref{fig:lattice-sensing}, we plot the distance traveled in this way versus the sensing range of the robot. Each data point is an average over 1000 trials. We consider exponential, geometric, Bernoulli and Pareto reward distributions. 

Notice that Figure~\ref{fig:lattice-sensing}(a)-(c) are semi-log plots, so the expected distance of travel scale exponentially with increasing sensing range for the light-tailed distributions. In other words, the sensing range is a logarithm of the distance traveled, as is stated by Theorem~\ref{theorem:lattice-sensing-light-tailed}. 

We run a slightly different experiment with Pareto-distributed rewards, where the robot stops if $\totalreward_i/m < \optimalmeanrewardtwo(m^{1.1}) - \delta$ and $m$ is the sensing range. This is due to the fact that, based on Theorem~\ref{theorem:lattice-optimal-heavy-tailed}, $\optimalmeanrewardtwo=\infty$ for all heavy-tailed rewards, so the distance-of-travel would have been 0. In Figure~\ref{fig:lattice-sensing}(d) shows the result, where the distance of travel increases only linearly (not exponentially) and is therefore much worse than the light-tailed distributions. This is consistent with our Theorem~\ref{theorem:lattice-sensing-pareto}. 

\subsection{Mean Reward in Continuous Spaces}

\begin{figure}[hbt]
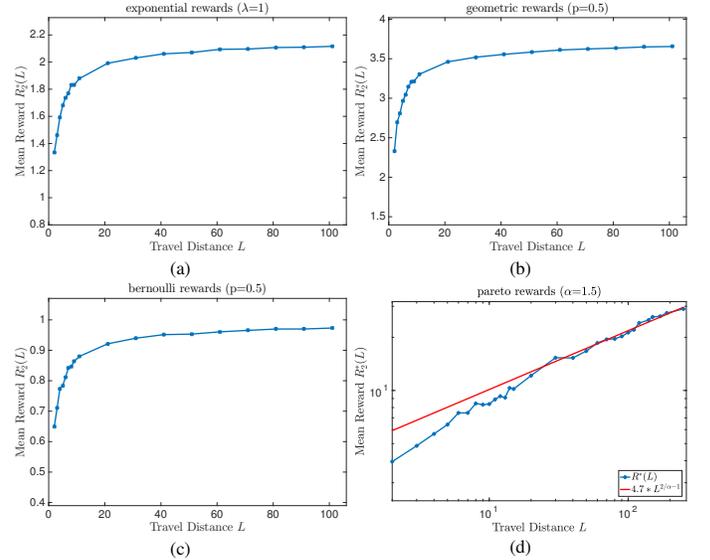

\myIncludeFourFigures
{data/reward_vs_length_cont/{reward_vs_length_cont.exponential.p=1.iter=1000}.eps}
{data/reward_vs_length_cont/{reward_vs_length_cont.geometric.p=0.5.iter=1000}.eps}
{data/reward_vs_length_cont/{reward_vs_length_cont.bernoulli.p=0.5.iter=1000}.eps}
{data/reward_vs_length_cont/{reward_vs_length_cont.pareto.p=1.5.iter=5000}.eps}
\caption{Optimal mean reward vs. travel distance on a two-dimensional poisson random reward field. The rewards associated with each target in the field are independent and identically distributed according to (a) exponential, (b) geometric, (c) Bernoulli and (d) Pareto distributions, respectively. (a)-(c) The mean reward for light-tailed distributions converges to a finite constant. (d) The mean reward for a heavy-tailed distribution goes to infinity.
}
\label{fig:cont-optimal-mean-reward}
\end{figure}

In this section, we verify Theorems~\ref{theorem:cont-limit-exists}-\ref{theorem:cont-optimal-heavy-tailed}, as well as Proposition~\ref{proposition:cont-optimal-pareto}, in Monte-Carlo simulations. We consider the continuous problem described in Section~\ref{section:problem:problem2}. The target locations are distributed according to a Poisson process with intensity $1$. We consider rewards that are distributed according to the exponential, geometric, Bernoulli and Pareto distributions. In Figure~\ref{fig:cont-optimal-mean-reward}, we plot the mean reward versus the travel distance for each of these reward distributions. 
As predicted by Theorems~\ref{theorem:cont-limit-exists}-\ref{theorem:cont-optimal-heavy-tailed}, as travel distance increases, the mean reward seems to converge towards a finite value for light-tailed distributions, as shown in Figure~\ref{fig:cont-optimal-mean-reward}(a)-(c). In Figure~\ref{fig:cont-optimal-mean-reward}(d), however, we show a log-log plot with Pareto rewards. It is clear that the optimal mean reward is not only diverging to infinity, but also growing at a rate of $O(L^{2/\alpha-1})$ as the travel distance $L$ increases, as predicted by Proposition~\ref{proposition:cont-optimal-pareto}.

\subsection{The Impact of Sensing Range On Performance}
In this section, we verify Theorem~\ref{theorem:cont-sensing-light-tailed}-\ref{theorem:cont-sensing-pareto} in Monte-Carlo simulations.
We consider the problem setup presented in Section~\ref{section:problem:problem2}. The target locations are distributed according to a Poisson process with intensity $\lambda=1$. The robot has limited sensing range, and it runs Algorithm~\ref{algorithm:continuous}. 
We fix a sensing range $S$, and we let the robot travel until the mean reward it collects goes under the value that is $\delta$ away from the optimal. We record the distance the robot can travel in this manner. 

\begin{figure}[hbt]
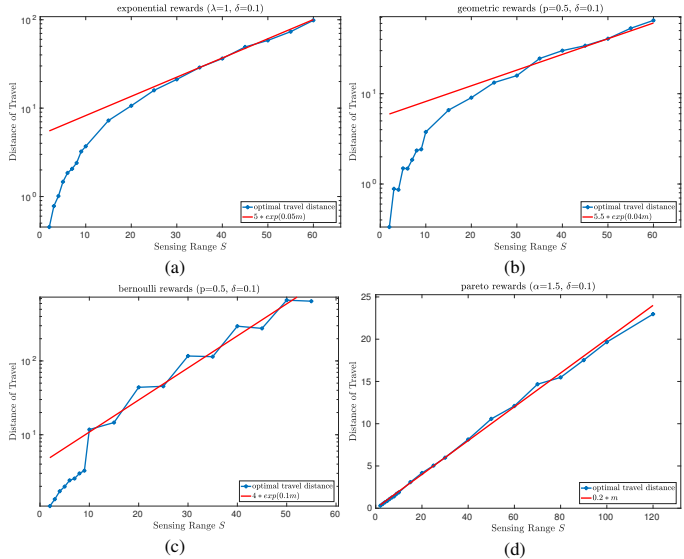


\myIncludeFourFigures
{data/dist_vs_horizon_cont/{dist_vs_horizon_cont.exponential.p=1.delta=0.1.iter=1000}.eps}
{data/dist_vs_horizon_cont/{dist_vs_horizon_cont.geometric.p=0.5.delta=0.1.iter=1000}.eps}
{data/dist_vs_horizon_cont/{dist_vs_horizon_cont.bernoulli.p=0.5.delta=0.1.iter=3000}.eps}
{data/dist_vs_horizon_cont/{dist_vs_horizon_cont.pareto.p=1.5.delta=0.1.iter=5000}.eps}

\caption{The average distance-of-travel is plotted against sensing distance $\sensingrange$ of the robot for the continuous problem. The Poisson process is parameterized with $\lambda = 1$. (a)-(c) are log-linear plots for exponential, geometric and bernoulli rewards, respectively. The distance-of-travel grows exponentially fast with sensing range when reward distribution is in the light-tailed family. (d) is a linear plot for pareto rewards (in the heavy-tailed family), and the distance-of-travel only grows linearly with sensing range.}

\label{fig:cont-sensing}
\end{figure}

In Figures~\ref{fig:cont-sensing}(a)-(c), we plot this distance traveled versus the sensing range in semi-log plots. We started the experiment with for exponentially distributed, geometric distributed and Bernoulli rewards, respectively. Notice that the distance traveled in this manner grows exponentially with increasing sensing range. Hence, in other words, the sensing range required to traverse a certain distance increases only logarithmically with the travel distance. The robot's performance, in terms of the reward collected, is still guaranteed to be a constant factor away from the optimal. 

In Figure~\ref{fig:cont-sensing}(d) we show the experiment with Pareto-distributed rewards. Similar to Section~\ref{sec:lattice-sensing}, we run a slightly different experiment where the baseline mean reward is $\optimalmeanrewardtwo(\sensingrange^{1.1})$ instead of $\optimalmeanrewardtwo$, where $\sensingrange$ is the sensing range. The optimal distance travelled is increasing only linearly with the sensing range $\sensingrange$.

\subsection{The Impact of Agility on Performance}

\begin{figure}[h]
\centering
\includegraphics[width=0.5\textwidth]{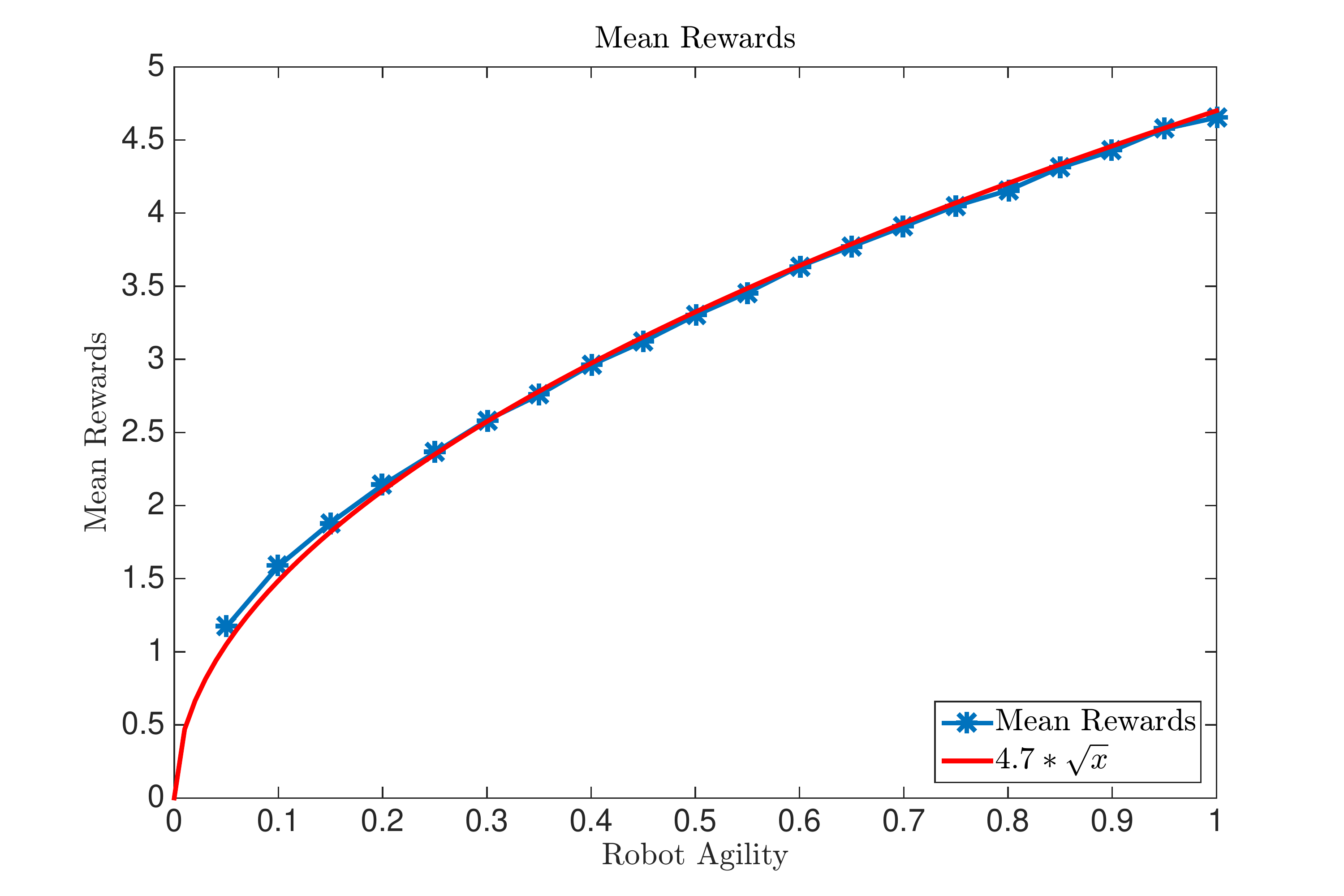}
\caption{Mean reward versus robot agility for the simulation where the intensity $\lambda$ of the Poisson point process is 10 and the reward follows an exponential distribution with mean 1.}
\label{fig:reward-vs-agility}
\end{figure}

In this section, we verify Theorem~\ref{theorem:agility} in Monte-Carlo simulations. 
We consider the problem setup described in Section~\ref{section:problem:problem2}. The locations of the targets are distributed according to a Poisson process with intensity $\lambda = 10$, and the rewards are exponentially distributed with unit mean. The travel distance is fixed to $L=30$. In these simulations, we vary the agility parameter and we observe how the mean reward varies. The results are shown in Figure~\ref{fig:reward-vs-agility}. Each data point is averaged over 300 independent trials. 

Notice that the mean reward follows the $\sqrt{A}$ rule, where $A$ is the agility of the robot, as stated by Theorem~\ref{theorem:agility}.

\subsection{Requirements on Computation}

\begin{figure}[hbt]
\myIncludeTwoFigures{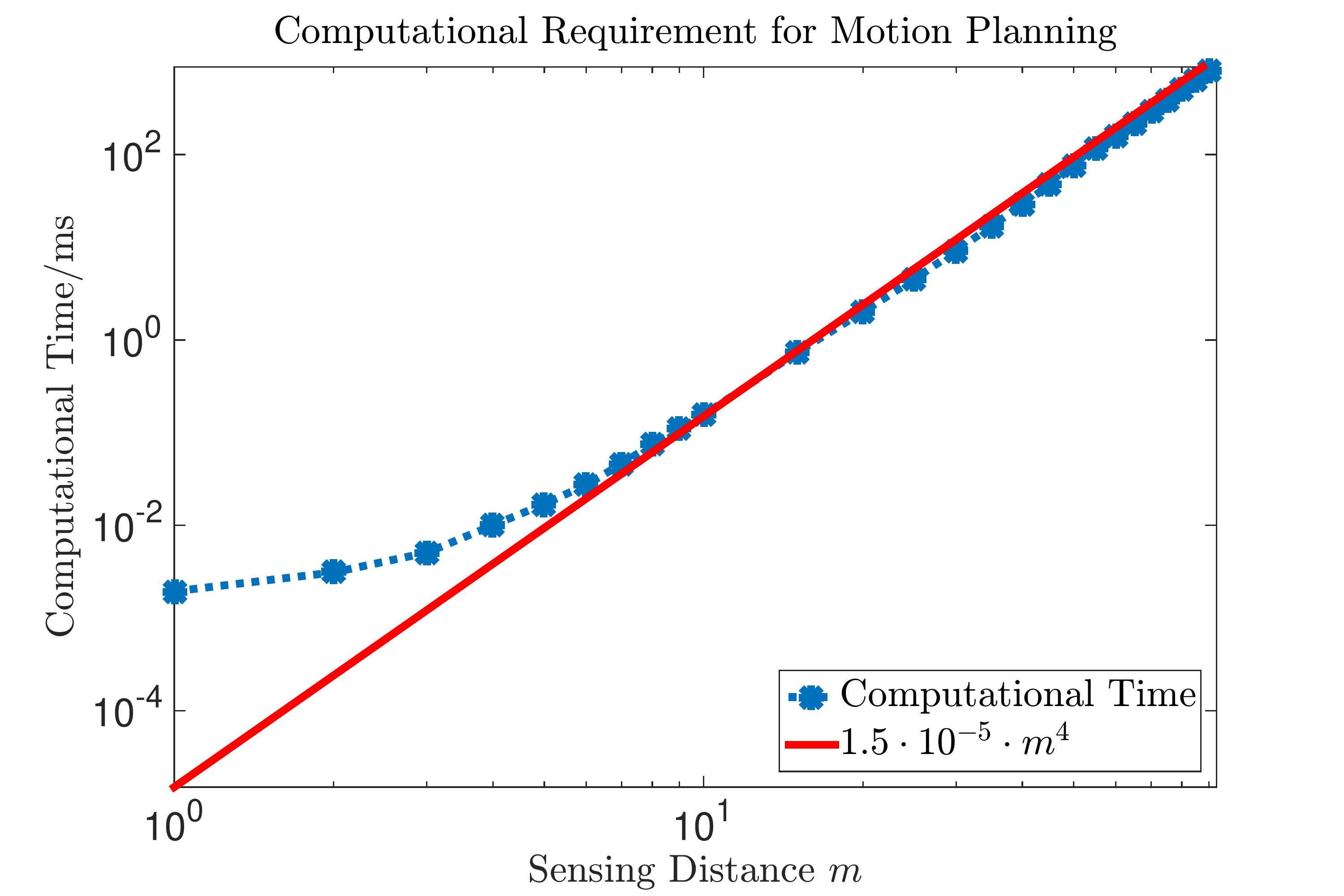}{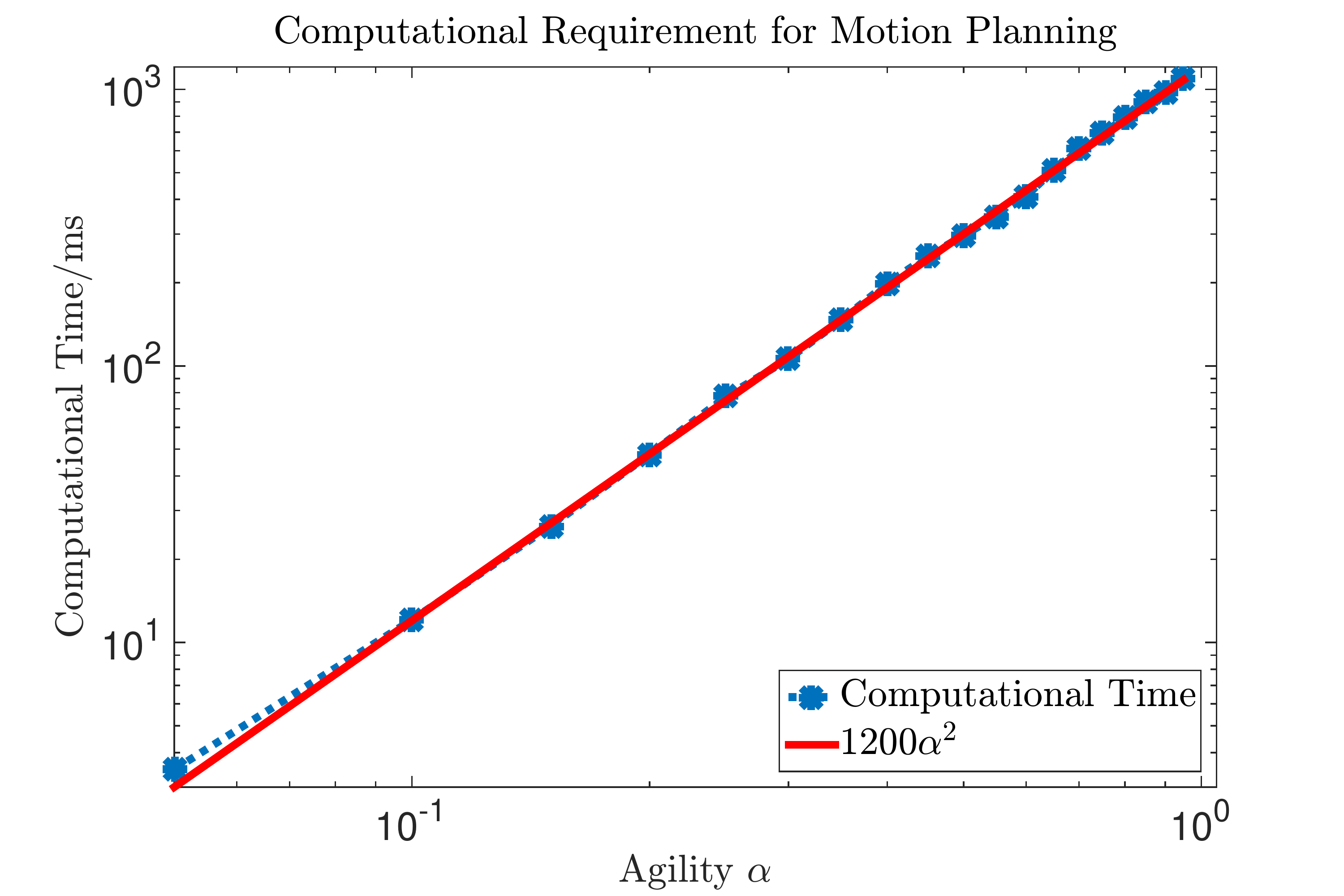}{1.1cm}{0}{2.5cm}{0}
\caption{Mean runtime of dynamic programming for motion planning versus sensing range $m$ and robot agility $\alpha$. The intensity of the Poisson point process is $\lambda=1$ and the reward follows a bernoulli distribution with $p = 0.5$.}
\label{fig:computation-planning}
\end{figure}

\begin{figure}[hbt]
\myIncludeTwoFigures{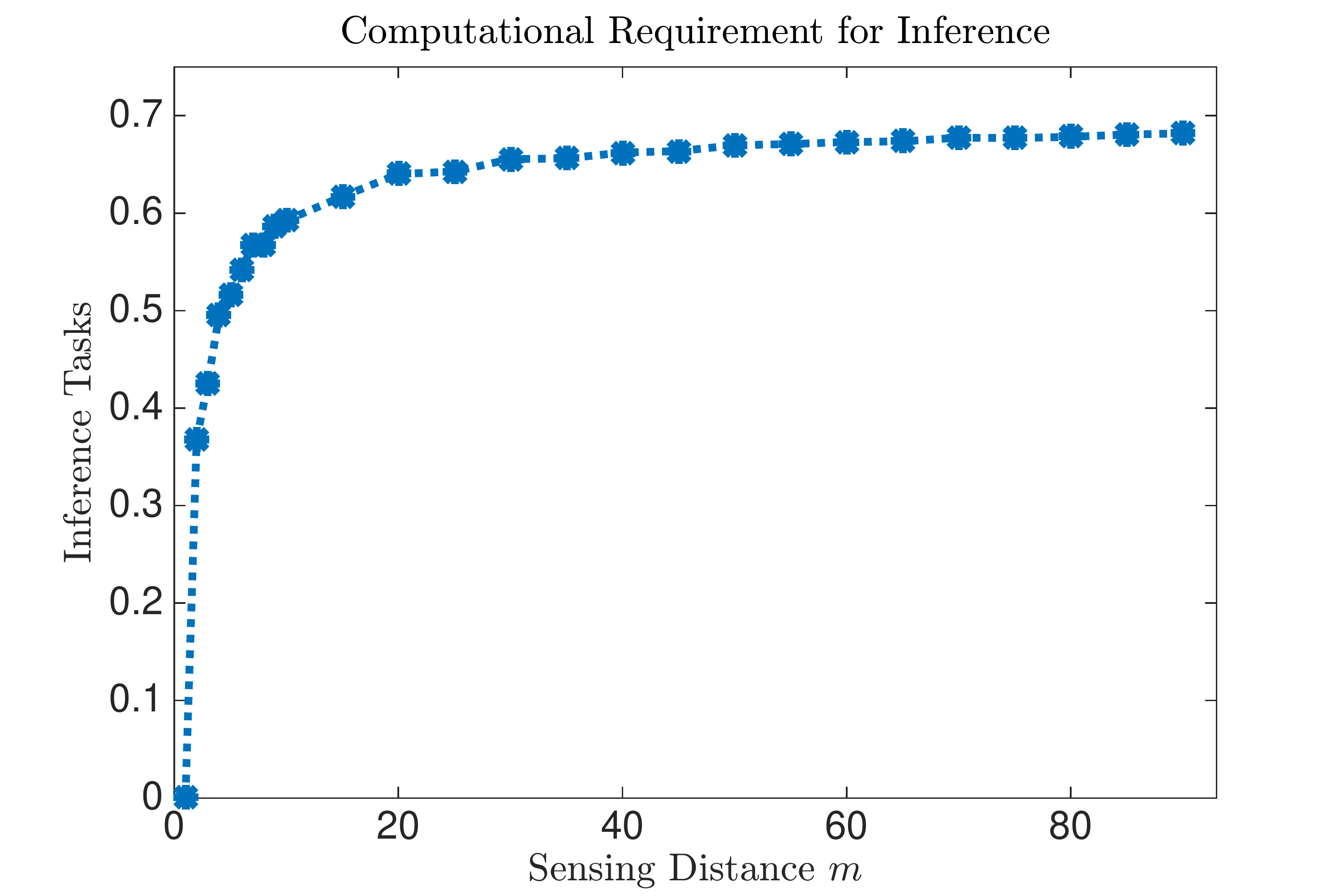}{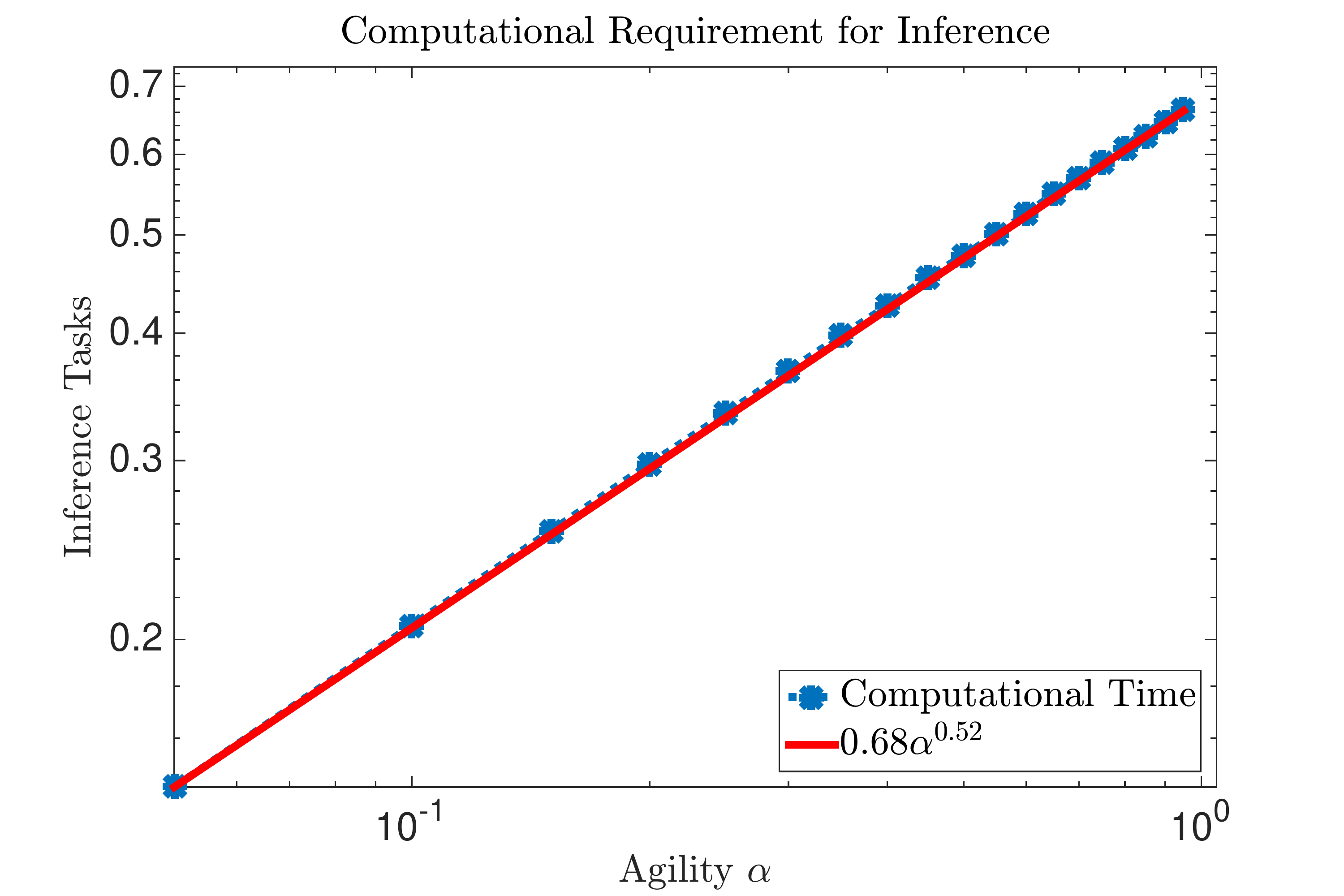}{1.1cm} {0} {2.5cm} {0}
\caption{Average number of inference tasks (targets visited) versus sensing range $m$ and robot agility $\alpha$. The number of targets visited grows with the robot's sensing distance but quickly reaches a plateau, and is linear with robot agility $\alpha$.}
\label{fig:computation-inference}
\end{figure}

In this section, we verify our claims in Section~\ref{section:computational_workload} in Monte-Carlo simulations. We consider the setting of the problem presented in Section~\ref{section:problem:problem2}. The robot travels with limited perception range, and we take a look at how sensing range and agility impacts the computation time devoted to planning and inference tasks. 
The results are presented in Figure~\ref{fig:computation-planning} for computation time devoted to motion planning, and in Figure~\ref{fig:computation-inference} for computation time devoted to inference. 

In Figure~\ref{fig:computation-planning}(a), we find that the computation time for motion planning scales as $\sensingrange^4$, where $\sensingrange$ is the sensing range, each time the algorithm is run. Hence, the computation required for motion planning is $\sensingrange^3$ per unit distance. 
In Figure~\ref{fig:computation-planning}(b), we observe that the computation time increases quadratically with increasing values of the agility parameter $\agility$. Notice that the computation time devoted to motion planning scales quadratically with increasing agility, as predicted in Section~\ref{section:computational_workload}. 

In Figure~\ref{fig:computation-inference}(a), we observe that the computation time devoted to inference (as measured by the number of inference tasks) is constant with increasing sensing range. In Figure~\ref{fig:computation-inference}(b), we see that the computation time devoted to inference grows roughly as $\sqrt{\agility}$, where $\agility$ is the agility of the robot, as predicted in Section~\ref{section:computational_workload}. 

%% file: discussion.tex


\section{Discussion}
\label{section:application}

In this section, we provide a brief discussion on our theoretical results, and illustrate how they can be used to gain valuable insight into design problems involving autonomous vehicles utilized in data gathering applications. First, in Section~\ref{section:codesign}, we outline the insights that our major findings presented in Section~\ref{section:analysis} provide. Second, in Section~\ref{section:sensor-selection}, we consider a sensor selection problem involving an aerial vehicle gathering data from unattended ground sensors, where we analyze sensing with homogeneous versus heterogeneous sensors.

\subsection{Insight for the Design of Sensing, Agility, and Computation Properties of Data-gathering Vehicles}
\label{section:codesign}

In this section, we discuss our major results in the context of the data-gathering problem presented in Section~\ref{section:problem:data_gathering} from the perspective of sensor precision, sensing range, agility, and computation capabilities. 
%
%
The main objective of this section is to establish the connections between the data-gathering problem presented in Section~\ref{section:problem:data_gathering} and the maximum-reward motion problem we presented in \ref{section:problem:problem2} and analyzed in Section~\ref{section:analysis}. 

\paragraph{On sensor precision} 
Recall from the data-gathering problem that the sensing precision $\beta$ corresponds to the reward in the maximum-reward motion problem. Hence, the larger the sensor precision $\beta$ in the former, the higher the corresponding reward in the latter.  
Then, according to Theorem~\ref{theorem:cont-limit-exists}, the precision of the estimate of the unknown variable $\theta$ will increase linearly with travel distance $L$, almost surely. Furthermore, the total precision of the estimate is at least $L \cdot \sqrt{\lambda \EE[\beta^2]}$, where $\lambda$ is the intensity of the target locations and $\beta$ is a random variable (either geometric or exponential) that denotes the precision of each measurement. 
%



Hence, we find that both increasing measurement precision ($\beta$) and increasing travel distance ($L$) has non-diminishing returns for the data-gathering problem outlined in Section~\ref{section:problem:data_gathering}. 

\paragraph{On sensing range}
The sensing range has different implications depending on the distribution of the precision of each measurement. 
If the precision of the measurements are light-tailed, then the precision of the estimate of $\theta$ increases linearly with increasing distance. Furthermore, according to Theorem~\ref{theorem:cont-sensing-light-tailed}, even with a sensing distance of $\log(L)$, the precision of the estimate of $\theta$ is almost as good as the precision of the same estimate when the sensing distance is $L$, if the vehicle travels a distance of $L$, with high probability. In other words, it is possible to achieve near-optimal estimation performance with little sensing distance for light-tailed reward.  

However, when the precision of the measurements is distributed according to the Pareto distribution with parameter $\alpha \in (0,2)$, then the precision of the estimate increases super-linearly with increasing sensing distance. Furthermore, according to Theorem~\ref{theorem:cont-sensing-pareto}, it is impossible to obtain near-optimal estimation performance with small sensing distance for Pareto reward with parameter $\alpha \in (0,2)$, which is a heavy-tailed distribution. We conjecture that this result applies to all heavy tailed distributions of the precision of the measurements. 

\paragraph{On agility}
According to Theorem~\ref{theorem:agility}, the precision of the estimate increases with increasing agility $\agility = w/v$, where $w$ is the maximum lateral speed and $v$ is the longitudinal speed of the vehicle (see Section~\ref{section:problem:problem2}). However, the increase comes with diminishing returns, proportional to $\sqrt{\agility}$.

\paragraph{On computation workload}
The computational workload is determined with the sensing range and the agility of the robot. 
The quantification of the computational workload for the data-gathering problem of Section~\ref{section:problem:data_gathering} follows that of the maximum-reward problem, the analysis for which was presented in Section~\ref{section:computational_workload}. 

The computational workload can be partitioned into two activities, namely motion planning and inference. The motion planning task consists of determining the set of target locations to be visited each time a new target gets in the sensing distance of the vehicle. The computational workload for this task increases substantially with increasing sensing distance and robot agility. Specifically, the the computational workload for planning increases as $O(\agility^2 \sensingrange^3)$ per unit time, where $\sensingrange$ is the sensing distance and $\agility$ is the robot agility. 

The inference task consists of incorporating the new measurements to improve the estimate. This task may be computationally challenging as it may involve image analysis, sensor fusion, {\em et cetera}. The computational workload for this task increases proportionally with $\sqrt{\agility}$, where $\agility$ is the agility of the robot. It is independent of the sensing range.

\hide{
\paragraph{The co-design of sensing distance, agility, and computation properties}
Finally, we note that in many problems, the sensing, agility, and computational capabilities of the vehicles must be designed jointly. 
For the purpose of illustration, we present a plot of mean reward versus both the sensing distance and the robot agility in Figure~\ref{fig:reward-vs-perception-agility} for exponentially-distributed reward, obtained from computational experiments. For a given sensing distance and agility, this plot shows how the reward per unit length (thus estimation precision per unit length) changes. From plots such as this one, sensing distance and agility can be determined for a desired level of perception. Let us note that some of the information on this plot can be proved in a mathematically rigorous manner as well, at least for large values of the sensing distance.   
Similar results are readily available for the computational workload as well. 


\begin{figure}[htbp]
\centering
\includegraphics[width=0.5\textwidth]{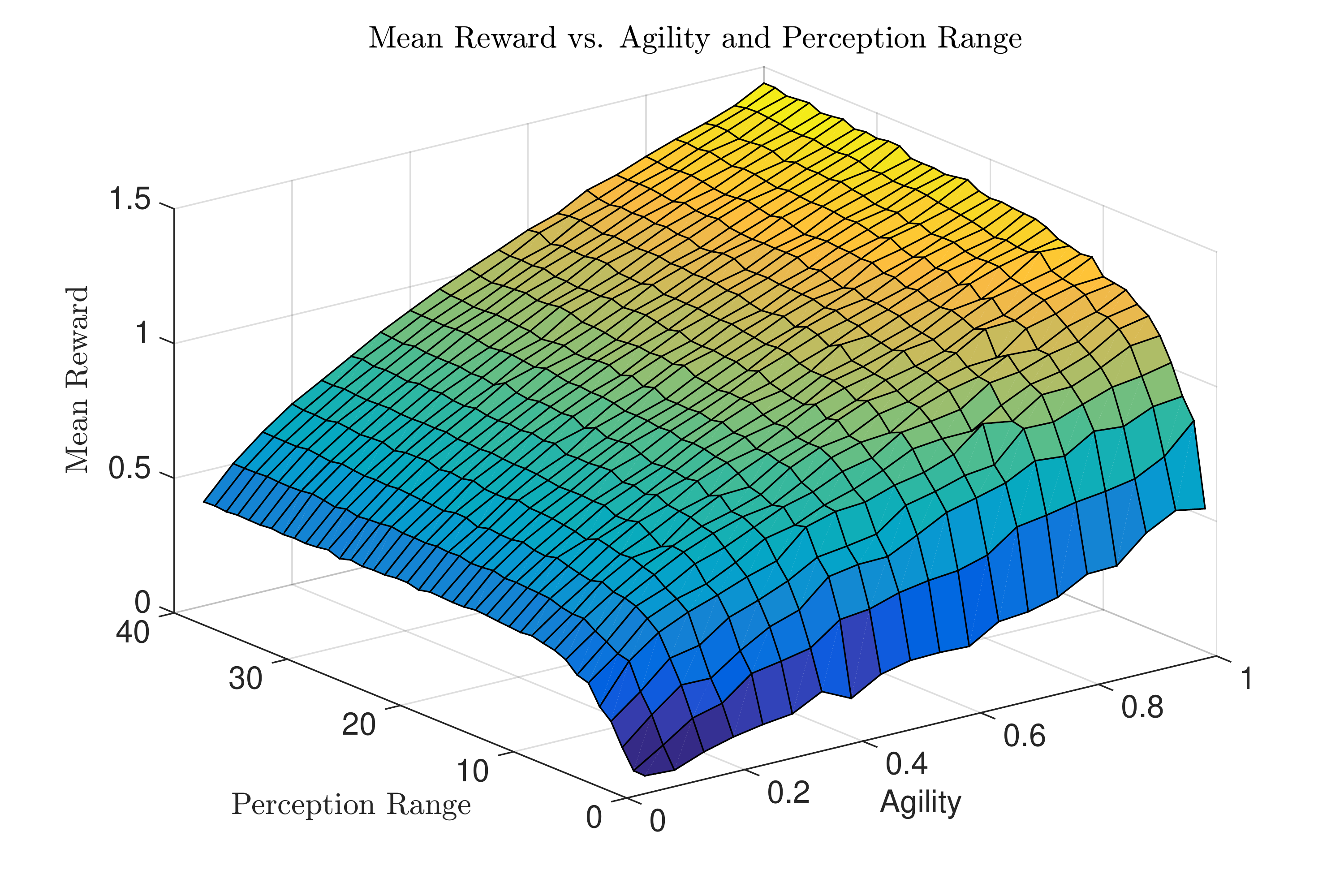}
\caption{Expected reward versus both the sensing distance and the agility for the simulation where the intensity $\lambda$ of the Poisson point process is 1 and the reward follows an exponential distribution with mean 1.}
\label{fig:reward-vs-perception-agility}
\end{figure}
}


\subsection{Case Study: Unattended Ground Sensor Selection}
\label{section:sensor-selection}

In this section, we present a short case study that involves a UAV is tasked with estimating an unknown variable $\theta$ with data acquired from Unattended Ground Sensors (UGS) that are randomly distributed over a region of interest. 

The UGS technology is an emerging technology that may have substantial impact in environmental monitoring, surveillance, and reconnaissance. The UGS often house primitive sensors that record various measurements, \eg, seismic, acoustic, magnetic, temperature, and humidity measurements, continuously for extended time periods, \eg, for several months. They are often deployed sparsely, which prevents formation of ad-hoc networks. However, the data they record can be collected by UAVs that fly over the sensors. 

In this section, we demonstrate how our analysis can be utilized to arrive at fundamental results for a certain kind of UGS selection problem. Specifically, we consider a problem where each UGS provides a measurement of the hidden variable $\theta$ corrupted with Gaussian noise. The precision of the measurement may depend on the quality of the UGS. We assume that the UAV recognizes each UGS from a certain distance, and learns the precision of the measurement that is obtained by that UGS. The UAV must plan its path carefully to best estimate the unknown variable $\theta$. Clearly, this problem is the same as the problem which we presented in Section~\ref{section:problem:data_gathering}.

Suppose we have the option to choose the sensors before they are distributed in the field. Due to limited budgets, the average quality of sensors is fixed and the total number of sensors is given, \ie, 
$$\E[\beta_i] = \mu_\beta,$$ 
where $\mu_\beta$ is some positive constant and $\lambda$ is known. With above constraints, we would like to address the following question: {\it Which one of the following two strategies yields a higher level of confidence for the estimation?
\begin{enumerate}
\item Assign the same level of precision to all sensors, \ie, $\beta_i = \mu_\beta$ for all $i$
\item Randomize the level of precision $\beta_i$ over some probability distributions $F_\beta$ with mean $\mu_\beta$
\end{enumerate} 
}

In the former option, all sensors are of the same quality; in the latter one, some sensors provide more precise (less noisy) measurements, while some others provide less provide (more noisy) measurements, when compared to the former option.

Notice that this sensor selection problem is an instance of the maximum-reward motion problem presented in Section~\ref{section:problem:problem2}. In this case, the reward is the precision $\beta_i$ of the measurement that the $i$th UGS provides, hence the quality of that UGS.

Let us analyze the performance of each of the strategies by computing the total reward collected in each case. As we established in the previous section, the total reward is proportional to the precision of our estimate of the hidden variable $\theta$ by visiting the unattended ground sensors.

The first strategy assigns equal precision to all UGS sensors. For optimal performance, the robot should visit as many sensors as possible in order to maximize the total precision gain. 
The resulting performance is analyzed below. 
\begin{theorem}[Adapted From \cite{seppalainen1997increasing}]
Suppose the reward locations are generated by a Poisson point process with intensity $\lambda$ on $\reals^2$ and all reward is 1. The robot dynamics satisfies the following ordinary differential equation:
$$
\dot{x}_1(t)  = v, \quad \dot{x}_2(t)  = u(t),
$$
where $\vert u(t) \vert \leq v$ (\ie, the robot agility is 1). Then
\begin{align*} 
\lim_{L \rightarrow \infty} \frac {\totalreward (L)} {L} = \sqrt{2\lambda} \quad \text{almost surely}.
\end{align*}
\label{theorem:5-2}
\end{theorem}
Theorem~\ref{theorem:5-2} provides the expected optimal mean reward collected when the rewards are equal to 1 surely. Therefore, it follows that the average number of sensors visited is $\sqrt{2\lambda}$, and hence the overall precision gain would be $L \cdot \sqrt{2\lambda} \cdot \E[\beta]$. 

The second strategy, on the other hand, utilizes sensors with random precisions (for instance, when the precisions follow an exponential distribution with mean equal to the homogeneous strategy). The exact $\optimalmeanrewardtwo$ for the Poisson random reward field is out of reach at the moment, but the computational experiments in Figure~\ref{fig:lattice-sensing}(a) show that when the mean $\E[\beta] = 1$, the expected precision gain is at least 2.1, much higher than the expected precision gain $\sqrt{2}$ when using homogeneous sensors. By the comparison of light-tailed (bounded variance) and heavy-tailed (infinite variance) distributions, we conjecture that the higher variance of the distributions, the better performance we can expect from the deployment of random-precision sensors.


\begin{figure}[htbp]
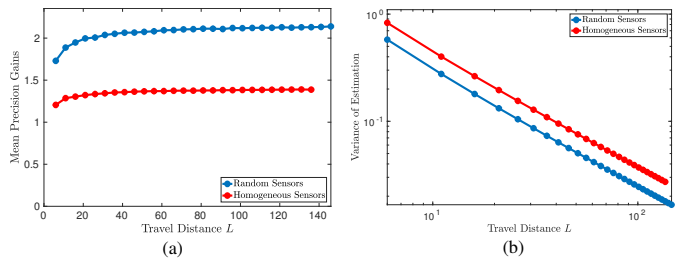

\centering
\myIncludeTwoFigures{{data/sensor/gains}.eps}{{data/sensor/variance}.eps}{0} {0} {0} {0}
\caption{Comparison of randomized sensors against homogeneous sensors by the mean precision gains and the variance of estimation. The randomized strategy assumes that sensor precisions follow an exponential distribution with mean 1, while the homogeneous sensors have a constant precision of $1$. The randomized strategy provides higher mean precision gain with lower variance.} 
\label{fig:ugs}
\end{figure}

To illustrate this comparison, we present results of a Monte-Carlo study. 
In Figure~\ref{fig:ugs}(a), we show the precision gains, and in Figure~\ref{fig:ugs}(b) we show how the variance of estimate (\ie, inverse of the precision gains) of the hidden variable decays. 
We observe that, when the sensor quality is randomized, the quality of the estimate is better and the variance of estimation decreases faster. 

%% file: appendices/proof-lattice-limit-exists.tex

\section{Proof for Theorem~\ref{theorem:lattice-limit-exists}}
\label{app:lattice-limit-exists}

Let's first introduce the definition of \emph{subadditivity} and the \emph{Fekete's Subadditive Lemma}.
\begin{definition}[Subadditivity~\cite{Steele:1996wn}] 
\label{definition:subadditivity}
A sequence $\{ a_n\}, n \geq 1$, is called \emph{subadditive} if it satisfies the inequality
$$
a_{n+m} \leq a_n + a_m.
$$
A sequence is \emph{superadditive} if 
$$
a_{n+m} \geq a_n + a_m.
$$
\end{definition}
\begin{lemma}[Fekete's Subadditive Lemma~\cite{Steele:1996wn}] 
\label{lemma:Fekete}
For every subadditive sequence $\{ a_n \}_{n=1}^\infty$, the limit $\lim_{n \to \infty} \frac {a_n} {n}$ exists and is equal to $\inf \frac {a_n} n$.
\end{lemma}

Now we return to the proof for Theorem~\ref{theorem:lattice-limit-exists}.
\begin{proof}
From Proposition 2.1 in~\cite{martin2006last} we learn that $\EE[\optimaltotalreward(n)]$ is superadditive, i.e., the sequence $-\EE[\optimaltotalreward(n)]$ is subadditive. Then the result follows directly from Lemma~\ref{lemma:Fekete}. 
\end{proof}

%% file: appendices/proof-lattice-optimal-light-tailed.tex


\section{Proof for Theorem~\ref{theorem:lattice-optimal-light-tailed}}
\label{app:lattice-optimal-light-tailed}
The problem of computing the maximum reward that the optimal path on a lattice can possibly collect is called the {\em last passage percolation problem}~\cite{martin2006last} in statistical mechanics, and it has connections with non-equilibrium statistical mechanics problems involving corner growth~\cite{johansson2001random}. 
This problem has attracted tremendous attention in the past. In particular, the function $\optimaltotalreward (v_\mathrm{dest})$ has been analyzed extensively.
%
Let us recall some of the main results from this literature and subsequently state and prove our main results for this section.
\begin{definition}[Shape function] 
\label{def-shape-function}
The function $g(\cdot)$, defined as
$$
g(\vertex) := \sup_{k \in \naturals} \frac{\EE[\optimaltotalreward (\floor{k\,\vertex})]}{k},
$$
is called the shape function.
\end{definition}
Firstly, the maximum reward towards any destination $v$ has been shown to converge to a limit, when reward is independent and identically distributed. 
\begin{proposition}[See Proposition 2.1 in~\cite{martin2006last}] 
\label{proposition:almost_sure_convergence}
Assume the reward $r(v)$ at each vertex $\vertex$ is an i.i.d. random variable and $\EE[r(\vertex)] < \infty$. Then, $\frac{\optimaltotalreward (\floor{k\,\vertex})}{k}$ converges to the shape function $g(\vertex)$ almost surely as $k$ diverges to infinity, \ie, 
$$
\PP \Big( \, \lim_{k \to \infty} \frac{\optimaltotalreward (\floor{k\,\vertex})}{k} = g(\vertex) \,\Big) = 1,
$$
and for all $\alpha > 0$, 
$$
\alpha g(\vertex) = g (\alpha \vertex).
$$
\end{proposition}
That is, for any $\vertex$, the maximum reward $\optimaltotalreward(\vertex)$ converges to the shape function $g(\vertex)$, almost surely. However, this limit may be infinite, depending on the distribution of $r(\vertex)$. Martin~\cite{martin2006last} provides a almost optimal necessary condition for the finiteness of the shape function $g$ as follows. 
\begin{theorem}[See Theorem 4.1 in~\cite{martin2006last}]
\label{theorem:finite-g}
If the reward distribution $F$ satisfies Equation~\ref{eqn:light-tail-reward-definition}, i.e.,
$$
\int_0^\infty ( 1 - F(s) )^{1/d} ds < \infty,
$$
then $g(\vertex) < \infty$ for all $\vertex \in \reals_+^d$. 
\end{theorem}
Now we proceed to the proof for Theorem~\ref{theorem:lattice-optimal-light-tailed}.
\begin{proof}
By the superadditivity of $\EE[\optimaltotalreward(n)]$ we have 
$$\EE[\optimaltotalreward(n\mathbf{1})] \leq \EE[\optimaltotalreward(nd)] \leq \EE[\optimaltotalreward(2nd\mathbf{1}) / 2],$$ 
where $\mathbf{1} = (1, 1, \dots, 1)$. If we divide the above inequalities all by $nd$, then for large $n$ both the left-hand-side and right-hand-side converge to the same constant due to the convergence result in Proposition~\ref{proposition:almost_sure_convergence}. It immediately follows that 
$$\lim_{n \to \infty} \EE \left[ \frac {\optimaltotalreward(nd)} {nd} \right] 
= \lim_{n \to \infty}\EE \left[\frac {\optimaltotalreward(n\mathbf{1})} {nd} \right] 
= \frac {g(\mathbf{1})} d .$$ 
Therefore, if Equation~\ref{eqn:light-tail-reward-definition} holds, we have
\begin{equation}
\label{eqn:r-star-is-g}
\expectedoptimalmeanreward_d 
= \lim_{n\to\infty}\EE[\optimalmeanreward(n)]  
= \lim_{n \to \infty} \EE \left[ \frac{\optimaltotalreward(n)}{n} \right ] 
= \frac {g(\mathbf{1})} d, 
\end{equation}
which is a finite constant by Theorem~\ref{theorem:finite-g}.
\end{proof}

%% file: appendices/proof-lattice-exponential-geometric.tex


\section{Proof for Proposition~\ref{proposition:lattice-exponential-geometric}}
\label{app:lattice-exponential-geometric}
The results in \cite{martin2006last} show that the shape function $g(v)$ can be computed exactly for at least two cases, namely, when the distribution $F$ of reward $r(v)$ at each vertex is either an exponential distribution or a geometric distribution. More specifically, if the reward distribution $F$ is an exponential distribution with parameter $\lambda = 1$, then the shape function $g$ defined in Proposition~\ref{proposition:almost_sure_convergence} is
\begin{equation}
\label{equation:4.2.1 exp}
g\big((\vertexnumber{1},\vertexnumber{2})\big) = (\sqrt{\vertexnumber{1}} + \sqrt{\vertexnumber{2}})^2, \quad \mbox{for all } (\vertexnumber{1},\vertexnumber{2}) \in \naturals^2.
\end{equation}
If the reward distribution $F$ is a geometric distribution with parameter $p$, \ie, $\PP(X = k) = p (1-p)^{k-1}$ for $k=1, 2, \dots$, then the shape function is
$$
g\big((\vertexnumber{1},\vertexnumber{2})\big) = \frac {\vertexnumber{1} + 2\sqrt{\vertexnumber{1}\vertexnumber{2}(1-p)} + \vertexnumber{2}} {p},  \, \mbox{for all } (\vertexnumber{1},\vertexnumber{2}) \in \naturals^2.
$$
Following Equation~\ref{eqn:r-star-is-g} in Appendix~\ref{app:lattice-optimal-light-tailed}, we readily derive that for both the geometric and exponential distributions
$$
\expectedoptimalmeanreward_2 = g(\mathbf{1})/2 = \mu + \sigma.
$$
Computing the shape function for other reward distributions, however, remains a long-standing, well-known open problem~\cite{martin2006last}. 

%% file: appendices/proof-lattice-optimal-pareto.tex

\section{Proof for Proposition~\ref{proposition:lattice-optimal-pareto}}
\label{app:lattice-optimal-pareto}
Let's first introduce a lemma that is useful for our proof.
\begin{lemma}[See Theorem 2.1 in~\cite{hambly2007heavy}] 
\label{lemma:discrete-heavy-help}
Suppose the CDF $F(x)$ is regularly varying with index $\alpha \in (0,2)$. Let $a_N = F^{-1}(1- 1/N)$, for all $N \in \naturals$. Then, $a_{n^2}^{-1}\, \totalreward(n)$ converges in distribution to a random variable $T$ that is almost surely finite.
\end{lemma}


%
Now, we return to the proof of Proposition~\ref{proposition:lattice-optimal-pareto}. For the Pareto distribution with parameters $x_m > 0$ and $\alpha \in (0,1)$, it is easy to derive that
\begin{equation*}
a_{n^2} = F^{-1}(1-1/n^2) = x_m \cdot n^{2/\alpha}
\end{equation*} 
By Lemma~\ref{lemma:discrete-heavy-help}, we have that $a_{n^2}^{-1}\, \optimaltotalrewardtwo(n)$ converges to a (non-trivial) random variable $T$ in distribution with $T < \infty$ almost surely. That is,
\begin{equation}
\frac {\optimaltotalrewardtwo(n)} {n^{2 / \alpha}} \rightarrow x_m \cdot T \quad \text{in distribution}
\end{equation} 
Note that $x_m \cdot T$ is almost surely bounded, which implies a finite expectation. It follows by the definition of $\optimalmeanrewardtwo$ that 
\begin{equation*}
\frac {\optimalmeanrewardtwo(n)} {n^{2 / \alpha - 1}} \rightarrow x_m \cdot T \quad \text{in distribution}.
\end{equation*} 

From Theorem~\ref{theorem:lattice-limit-exists} we know that $\optimalmeanrewardtwo(n) \to \optimalmeanrewardtwo$ surely. Therefore, it can be seen that as depending on the growth rate of $\optimalmeanrewardtwo(n)$ with respect to $n$, $\frac {\optimalmeanrewardtwo(n)} {n^{2 / \alpha - 1}}$ might converge to 0, a finite positive constant $c$, or $\infty$. 

By Skorokhod's Representation Theorem~\cite{billingsley2009convergence}, there exists a sequence of random variables $X_n$, defined on the same probability space, such that $X_n$ has the same distribution as $\frac {\optimalmeanrewardtwo(n)} {n^{2 / \alpha - 1}}$ and that $
X_n \rightarrow x_m \cdot T \quad \text{almost surely}$. Therefore we can apply Fatou's Lemma~\cite{royden1988real} here and obtain
\begin{equation*}
\liminf_{n \to \infty} \EE X_n \geq \EE \left[ x_m \cdot T \right] > 0
\end{equation*}
which eliminates the possibility that $X_n$ (and therefore 
$\frac {\optimalmeanrewardtwo(n)} {n^{2 / \alpha - 1}}$) converges to 0. In other words, $\frac {\optimalmeanrewardtwo(n)} {n^{2 / \alpha - 1}}$ must converge to either a finite constant $c > 0$ or $\infty$, and hence $\optimalmeanrewardtwo(n)$ is at least of growth rate $n^{2 / \alpha - 1}$. It immediately follows that
$$
\optimalmeanrewardtwo = \lim_{n \to \infty} \EE [\optimalmeanrewardtwo(n)] = \infty.
$$ 

%% file: appendices/proof-lattice-sensing-light-tailed.tex
\section{Proof for Theorem~\ref{theorem:lattice-sensing-light-tailed}}
\label{app:lattice-sensing-light-tailed}

To prove Theorem~\ref{theorem:lattice-sensing-light-tailed}, we adopt a two-step strategy: we first ``truncate'' the original problem with light-tailed reward to a simplified problem with bounded reward, and then apply concentration of measure to show that the difference between the two problems is actually small.

\subsection{``Truncated'' Problem with Bounded Reward}
Now let's first consider a truncated version of the original problem. Instead of having light-tailed distributed reward, we define a new problem where the reward associated with vertex $\vertex$ is now defined as $r^{(L)} = \max(r(\vertex), L)$, where $L >0$. We state the result for the truncated problem as follows.
\begin{lemma}\label{lemma:preliminaries:sensing_range_bounded}
Suppose the rewards $r(v)$ are independent, identically distributed and almost surely bounded.
Then, for any $\delta > 0$, there exists a constant $c$ such that $\IMPmeanreward(n,c\log n)$ converges to $\expectedoptimalmeanreward_d$ in probability, \ie, 
$$
\lim_{n \to \infty}\PP\left( \,\big\vert \IMPmeanreward(n,c\log n)  - \expectedoptimalmeanreward_d \big\vert \,\ge\, \delta \,\right) \,\,=\,\, 0.
$$
\end{lemma}

Before proving Lemma~\ref{lemma:preliminaries:sensing_range_bounded}, we state an intermediate result that enables our proof. This intermediate result is a concentration inequality, which plays a key role in deriving many results in nonequilibrium statistical mechanics~\cite{martin2002linear}.

\begin{lemma}[See \cite{martin2002linear}]
\label{lemma:4.2.3}
Let $\{Y_i,i \in \mathcal{I}\}$ be a finite collection of independent random variables that are bounded almost surely, \ie, $\PP(|Y_i| \leq L) = 1$ for all $i \in \mathcal{I}$. Let $\mathcal{C}$ be a collection of subsets of $\mathcal{I}$ with maximum cardinality $R$, \ie, $\max_{C \in \mathcal{C}}|C| \leq R$
and let $Z = \max_{C \in \mathcal{C}}\sum_{i \in C}Y_i.$
Then for any $u > 0$,
$$
\PP(|Z - \mathbb{E}Z| \geq u) \leq \exp\bigg( - \frac {u^2}{64RL^2} + 64\bigg).
$$
\end{lemma}
Now, we present the proof for Lemma~\ref{lemma:preliminaries:sensing_range_bounded}.
\begin{proof}[Proof for Lemma~\ref{lemma:preliminaries:sensing_range_bounded}]
First, note that:
\begin{align*}
& \lim_{n \rightarrow \infty}\PP\left(\big| \IMPmeanreward(n, m) - R_d^* \big| \geq \delta \right) \notag \\
=& \lim_{n \rightarrow \infty}\PP\left(\Big| \IMPmeanreward(n, m) - \frac {\mathbb{E}[\totalreward(n)]}{n} +  \frac {\mathbb{E}[ \totalreward(n)]}{n} - R_d^* \Big| \geq \delta \right) \notag \\
=& \lim_{n \rightarrow \infty} \PP\left(\Big| \IMPmeanreward(n, m) - \frac {\mathbb{E}[\totalreward(n)]}{n} \Big| \geq \frac \delta 2 \right) \\
&\qquad\qquad + \lim_{n \rightarrow \infty} \PP\left(\Big| \frac {\mathbb{E} [\totalreward(n)]}{n} - R_d^* \Big| \geq \frac \delta 2 \right) \notag 
\end{align*}
The second term is $0$ by Theorem~\ref{theorem:lattice-limit-exists}, so we focus on the first term. 
Let's define $\delta = \frac u n$. By the definition of $\IMPmeanreward(n,m)$: 
\begin{align}
& \lim_{n \rightarrow \infty}\PP\left(\big| \IMPmeanreward(n, m) - \optimalmeanreward \big| \geq \delta \right) \notag \\
=& \lim_{n \rightarrow \infty} \PP\left(\big| \IMPmeanreward(n, m) - \frac {\mathbb{E}\totalreward(n)}{n} \big| \geq \frac \delta 2 \right) \\
=& \lim_{n \rightarrow \infty} \PP\left( \big| \frac {\sum_{i=1}^{\frac n m}\totalreward_i(m)}{n} - \frac {\mathbb{E}\totalreward(n)}{n} \big| \geq \frac \delta 2  \right) \notag \\
=& \lim_{n \rightarrow \infty} \PP \left( \big| \frac {\sum_{i=1}^{\frac n m}\totalreward_i(m)}{n} - \frac {\mathbb{E}\totalreward(m)}{m} + \frac 	{\mathbb{E}\totalreward(m)}{m} - \frac {\mathbb{E}\totalreward(n)}{n} \big| \geq \frac \delta 2  \right) \label{eq:4.1-2}\\
\leq& \lim_{n \rightarrow \infty} \PP \bigg( \{\big| \frac {\sum_{i=1}^{\frac n m}\totalreward_i(m)}{n} - \frac {\mathbb{E}\totalreward(m)}{m} \big| \geq \frac \delta 4) \} \nonumber \\
	& \qquad \qquad  \bigcup \{ \frac {\mathbb{E}\totalreward(m)}{m} - \frac {\mathbb{E}\totalreward(n)}{n} \big| \geq \frac \delta 4 \} \bigg), \label{eq:4.1-3} \\
\leq& \lim_{n \rightarrow \infty} \PP \bigg( \big| \frac {\sum_{i=1}^{\frac n m}\totalreward_i(m)}{n} - \frac {\mathbb{E}\totalreward(m)}{m} \big| \geq \frac \delta 4 \bigg) \nonumber \\
	& \qquad \qquad + \lim_{n \rightarrow \infty} \PP \bigg( \big| \frac {\mathbb{E} \totalreward(m)}{m} - \frac {\mathbb{E} \totalreward(n)}{n} \big| \geq \frac \delta 4 \bigg) \label{eq:4.1-4} \\
\leq & \lim_{n \rightarrow \infty} \PP \bigg( \sum_{i=1}^{\frac n m}\big| \totalreward_i(m) - \mathbb{E} \totalreward(m)\big| \geq \frac {n\delta} 4 \bigg) \nonumber \\
  	& \qquad \qquad +  \lim_{n \rightarrow \infty} \PP \bigg(\big| \frac {\mathbb{E} \totalreward(m)}{m} - \frac {\mathbb{E} \totalreward(n)}{n} \big| \geq \frac \delta 4 \bigg).   \label{eq:4.1-5}
\end{align}
Again by Theorem~\ref{theorem:lattice-limit-exists}, both $\frac {\mathbb{E} \totalreward(m)}{m}$ and $\frac {\mathbb{E} \totalreward(n)}{n}$ converge to the same constant $R_d^*$ when $n$ and $m$ goes to infinity. Therefore, the second term vanishes. 
To help understand the proof, let's explain the inequalities above. The inequality between line \eqref{eq:4.1-2} and line \eqref{eq:4.1-3} can be derived by using the fact that
\begin{align*}
&\big\{\big|( \frac {\sum_{i=1}^{\frac n m} \totalreward_i(m)}{n} - \frac {\mathbb{E} \totalreward(m)}{m}) + (\frac {\mathbb{E} \totalreward(m)}{m} - \frac {\mathbb{E} \totalreward(n)}{n}) \big| \geq \frac \delta 2 \big\} \subset \\
&\big\{\big| \frac {\sum_{i=1}^{\frac n m} \totalreward_i(m)}{n} - \frac {\mathbb{E} \totalreward(m)}{m} \big | \geq \frac \delta 4 \big\} \bigcup \big\{ \big | \frac 	{\mathbb{E} \totalreward(m)}{m} - \frac {\mathbb{E} \totalreward(n)}{n} \big| \geq \frac \delta 4 \big\}
\end{align*}
Union bound is applied between between line \eqref{eq:4.1-3} and line \eqref{eq:4.1-4}. Line \eqref{eq:4.1-4} and line \eqref{eq:4.1-5} comes from the simple fact that 
$$\big| \frac {\sum_{i=1}^{\frac n m} \totalreward_i(m)}{n} - \frac {\mathbb{E} \totalreward(m)}{m} \big| \leq \sum_{i=1}^{\frac n m}\big| \totalreward_i(m) - \mathbb{E} \totalreward(m)\big|.$$ 

Now let's use Lemma~\ref{lemma:4.2.3}. Let $\mathcal{I}$ be the collection of nodes in the lattice. Define $\mathcal{C} = \{\mathcal{N}(\pi), \pi \in \Pi\}$, where  $\mathcal{N}(\pi) = \{\mathbf{v} \in \pi\}$ is the set of nodes in the path $\pi$.  Then, for the maximum-reward path with at most $n$ steps, the maximum cardinality is $\max_{C \in \mathcal{C}} |C| \leq n$. 
Then, by substituting $\totalreward(n)$ for $Z$ in Lemma~\ref{lemma:4.2.3}, 
$$
P(| \totalreward(n) - \mathbb{E}[\totalreward(n)] | \geq u) \,\,\leq\,\, \exp\Big(-\frac{u^2}{64nL^2} + 64\Big).
$$

Using the inequality we just derived and setting $m = c\log n$, we obtain
\begin{align}
 =& \lim_{n \rightarrow \infty}\PP \left( \left| \IMPmeanreward(n, m) - R_d^* \right| \geq \delta \right) \notag \\
 	\leq& \lim_{n \rightarrow \infty}  \PP \bigg(\sum_{i=1}^{n/m}\big|\, \totalreward_i(m) - \mathbb{E}[\totalreward(m)]\,\big| \geq \frac {n\delta} {4} \bigg)  \label{eq:4.1-6}\\
	\leq& \lim_{n \rightarrow \infty} \sum_{i=1}^{\frac n m} \PP \left( \big|\, \totalreward_i(m) - \mathbb{E}[\totalreward(m)] \,\big| \geq \frac {m\delta}{4} \right) \label{eq:4.1-8}\\
 	\leq& \lim_{n \rightarrow \infty} {\frac n m} \cdot\exp \left(-\frac{(\frac {m\delta} 4)^2}{64mL^2} + 64 \right)  \label{eq:4.1-9}\\
	=& \lim_{n \rightarrow \infty} {\frac 1 {c\, \log n}} \cdot \exp \left( \left(1 -\frac{ \delta^ 2}{1024 L^2} \cdot c \right) \log n + 64 \right) 
 	\label{eq:4.1-10}
\end{align}
The first inequality comes from line \eqref{eq:4.1-5}.  Union bound is again applied between between line \eqref{eq:4.1-6} and line \eqref{eq:4.1-8}. Lemma~\ref{lemma:4.2.3} is applied in line \eqref{eq:4.1-9}.  Line\eqref{eq:4.1-10} converges to 0 when the constant $c$ is sufficiently large, \ie, for any constant $c \geq \left( \frac {16L} {\delta} \right)^2$.
\end{proof}

\subsection{Proof for the Light-tailed Problem}
We proceed to prove results for the original problem with light-tailed reward (\ie, Theorem~\ref{theorem:lattice-sensing-light-tailed}) by applying concentration of measure. Let's first introduce a lemma.
\begin{lemma}[See \cite{martin2006last,talagrand1995concentration,martin2002linear}]
\label{lemma:truncation-bound}
Assume Equation~\ref{eqn:light-tail-reward-definition} holds true. Then there exists $c=c(d) < \infty$ such that for all $L>0$ and all $\bfx < \mathbf{1}$, the shape function defined in Proposition~\ref{proposition:almost_sure_convergence} satisfies
$$
g(\bfx) - g^{(L)} (\bfx) \leq c \int_L^\infty (1 - F(x))^{1/d} dx.
$$
\end{lemma}
Similar to the proof for Lemma~\ref{lemma:preliminaries:sensing_range_bounded}, we will be using union bounds to break down the terms in the original inequality. More specifically,
\begin{align*}
&  \PP \left( \left| \IMPmeanreward(n, m) - \expectedoptimalmeanreward_d \right| \geq \delta \right) \notag \\
=\,& \PP \big( \big| \IMPmeanreward(n, m) - \IMPmeanrewardTruncated(n, m) + \IMPmeanrewardTruncated(n, m) \notag \\
& \quad\quad\quad\quad - \expectedoptimalmeanreward_d + \optimalmeanrewardTruncated - \optimalmeanrewardTruncated \big| \geq \delta \big) \notag \\
\leq\,& \PP \left( \big| \IMPmeanreward(n, m) - \IMPmeanrewardTruncated(n, m) \big| \geq \delta/3 \right) \notag \\
& \quad + \PP \left( \big| \IMPmeanrewardTruncated(n, m) - \optimalmeanrewardTruncated \notag \big| \geq \delta/3 \right) \\
& \quad + \PP \left( \big| \expectedoptimalmeanreward_d - \optimalmeanrewardTruncated \big| \geq \delta/3 \right) \notag
\end{align*}
The first and the third terms are the gap between the truncated and the original problems. By Lemma~\ref{lemma:truncation-bound} both terms can be made arbitrarily small by choosing a large truncation threshold $L$. The second term also vanishes due to Lemma~\ref{lemma:preliminaries:sensing_range_bounded}, when the sensing range is of order $O(\log n)$, as $n$ increases. Therefore, with proper choice of the threshold $L$ and sensing range $m=O(\log n)$, 
$$
\lim_{n \to \infty} \PP \left( \left| \IMPmeanreward(n, m) - \expectedoptimalmeanreward_d \right| \geq \delta \right) = 0.
$$

%% file: appendices/proof-lattice-sensing-pareto.tex

\section{Proof for Theorem~\ref{theorem:lattice-sensing-pareto}}
\label{app:theorem-heavy-discrete}

Recall that $\IMPtotalreward(n;m)$ is defined as $\sum_{i = 1}^{n/m} \totalreward_i$ in Equation~\ref{eqn:definition-T-IMP}, where $\totalreward_i$ is the reward collected at the $i^{th}$ iteration in a receding horizon manner. We can apply the same argument in Appendix~\ref{app:agility} to each $\totalreward_i$ and derive that
\begin{equation*}
\totalreward_i = O(m^{2/\alpha}).
\end{equation*}
Therefore, we obtain
\begin{align*}
\IMPmeanreward \left( n; m \right) & = \frac {\sum_{i = 1}^{n/m} \totalreward_i(m)} {n} = O\left( m^{2/\alpha - 1} \right) 
\end{align*} 
Since we already have $\optimalmeanrewardtwo(n) = O(n^{2 / \alpha - 1})$ from Proposition~\ref{proposition:lattice-optimal-pareto}, it immediately follows that when $m = M(n)$ where $M(n)$ is a sub-linear function of $n$, we have
$$
\lim_{n \to \infty}\frac {\IMPmeanreward \left( n; M(n) \right)} {\optimalmeanrewardtwo(n)} = \lim_{n \to \infty} \left( \frac {M(n)} {n} \right)^{2/\alpha - 1} = 0
$$
%


%% file: appendices/proof-cont-limit-exists.tex

\section{Proof for Theorem~\ref{theorem:cont-limit-exists}}
\label{app:cont-limit-exists}

The proof applies a similar argument to Appendix~\ref{app:lattice-limit-exists}, using the lemma below (a continuous counterpart of Fekete's subadditive lemma). 
\begin{lemma}[See~\cite{hille1996functional}]
For every measurable subadditive function $f: (0,\infty) \to \reals$, the limit $\lim_{t \to \infty} \frac {f(t)} t$ exists and is equal to $\inf_{t>0} \frac {f(t)} t$.
\end{lemma}
The function $\EE [\optimaltotalreward(L)]$ is super-additive and by using the above lemma, it follows that the limit $\lim_{L \to \infty} \frac {\EE [\optimaltotalreward(L)]} L$ exists and is equal to $\sup_L \frac {\EE \optimaltotalreward(L)} L$.

%% file: appendices/proof-cont-optimal-heavy-tailed.tex
\section{Proof for Theorem~\ref{theorem:cont-optimal-heavy-tailed}}
\label{app:cont-optimal-heavy-tailed}
Let's denote the reachable region of the robot as $\mathcal{R}$. First note that 
$$\optimaltotalrewardtwo(L) \geq \max_{\vertex \in \mathcal{R}} r(\vertex),$$ 
for all $\vertex \in \mathcal{R}$. If $\EE r^d = \infty$, then for arbitrary $c>0$ and infinitely many $L$, there must exist $\vertex$ such that $r(\vertex) \geq cL$. In other words, there exists infinitely many $L$ such that for any $c$,
$$
\optimalmeanrewardtwo(L) 
= \frac {\optimaltotalrewardtwo(L)} L 
\geq \frac {\max_{\vertex \in \mathcal{R}} r(\vertex)} L 
\geq c
$$ 
Therefore, $\optimalmeanrewardtwo(L)$ does not concentrate, and thus $\expectedoptimalmeanreward_2 = \infty$.

%% file: appendices/proof-cont-optimal-pareto.tex

\section{Proof for Proposition~\ref{proposition:cont-optimal-pareto}}
\label{app:cont-optimal-pareto}

To analyze the continuous problem, we approximate the continuous reward field with a two-dimensional $N \times N$ regular lattice. This approximation turns the continuous problem into a discrete problem that we are already familiar with from previous discussions. 

More specifically, consider the set $\{ \frac 1 n*L, \frac 2 n *L, \dots, \frac {n-1} n *L, L \}^2 \subset [0, L]^2$. Let the node $\vertex = (i,j)$ on this discrete set be associated with $\rewardApproximate_{i,j}$, which is the total reward that falls within the region $[\frac i n *L, \frac {i+1} n * L) \times [\frac j n *L, \frac {j+1} n * L)$ in the continuous model. This regions has an area of $s = \left( \frac L N \right)^2.$

Since the reward locations are distributed randomly in the field according to a Poisson process with parameter $\lambda$, the number of targets within any of such regions follows a Poisson distribution with intensity $p = \lambda \cdot s =  \lambda \left( \frac L N \right)^2$. If we denote number of rewards within this region with a random variable $K \sim \poisson(p)$, then the total reward is 
$$\rewardApproximate_{i,j} = \sum_{k=1}^K r_k,$$ 
where $r_k$ is drawn i.i.d. from the common reward distribution $F$.

Note that on this discrete set, all previous results (Theorems~\ref{theorem:lattice-limit-exists}-\ref{theorem:lattice-sensing-pareto}) apply. It remains to show that the discrete set is indeed a good approximate of the continuous model, and in fact this has been given in the proof for Theorem 2.1 in~\cite{Hambly:2007wm} for Pareto distributions. We rephrase the result in our notation as follows:
\begin{theorem}
$\optimaltotalrewardtwo(1)$ is almost surely finite and as $n \to \infty$
$$
\frac {\optimaltotalrewardlattice{n}} {n^{2/\alpha}} \rightarrow \optimaltotalrewardtwo(1) \quad \mbox{in distribution},
$$
where $\optimaltotalrewardlattice{n}$ is the total reward on the approximate lattice.
\end{theorem}
Again, by the Skorohod Representation Theorem, we can define all the variables on the
same probability space in such a way that the convergence occurs almost surely. 

It is easy to see that for arbitrary size $L>0$, the optimal reward on the continuous space can be approximated by $\optimaltotalrewardlattice{Ln}$. Therefore by Proposition~\ref{proposition:cont-optimal-pareto} we have 
$$
\optimaltotalrewardtwo(L) = O(L^{2/\alpha-1}).
$$

%% file: appendices/proof-cont-agility.tex
\section{Proof for Theorem~\ref{theorem:agility}}
\label{app:agility}

\begin{proof}
When $\alpha = 1$, the robot is operating on an isosceles right triangle region, as shown in Figure~\ref{figure:right-angled-triangle}. However, if $\alpha \not= 1$, the robot is operating on some isosceles triangle with the vertex angle being $2\arctan \alpha$, as shown in Figure~\ref{figure:isosceles}. 
\begin{figure}[htbp]
\centering
	\begin{subfigure}[b]{0.45\linewidth}
        	        \centering
                \includegraphics{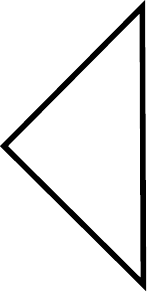}
                \caption{$\text{agility } \alpha = 1$}
                \label{figure:right-angled-triangle}
        \end{subfigure}
        \begin{subfigure}[b]{0.45\linewidth}
        	        \centering
                \includegraphics{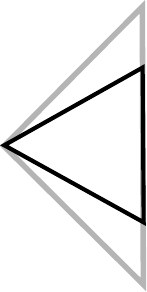}
                \caption{$\text{agility } \alpha \not= 1$}
                \label{figure:isosceles}
        \end{subfigure}
        \caption{The reachable set of the robot with drift, whose dynamics is described mathematically by Equations~\ref{equation:system}.}
\end{figure}

We extend all previous results to this isosceles triangle region by using the invariance property of Poisson point processes. By undergoing a deterministic transformation of the Poisson point process, we show the equivalence between this new problem (with non-unit $\alpha$) and the one with $\alpha = 1$. 

More specifically, this transformation is characterized by the mapping theorem of Poisson point process in \cite{streit2010Poisson}. Formally, let $f:\reals^2 \rightarrow \reals^2$ be the transformation between two poisson point processes. If $f$ is an affine transformation 
$$f(x) = Ax + b,$$ 
with $A \in \reals^{2 \times 2}$ being invertible, then the transformed Poisson Poisson has the new intensity
$$
\nu (y) = \frac {1} {|A|} \lambda \left(f^{-1}(y) \right) \left(A^{-1} (y - b) \right),
$$  
where $\lambda$ is the intensity of the original poisson point process. 

In this case, the transformation from an isosceles triangle to an isosceles right triangle is linear.  Therefore, $f = Ax$, where
$$
A = \begin{bmatrix}
       1 & 0 \\[0.3em]
       0 & 1/ \alpha        
     \end{bmatrix}.
$$
Geometrically, this transformation stretches (or compresses) the isosceles triangle vertically until the vertex angle is $90$ degrees, as shown in Figure~\ref{figure:isosceles}. Note that we assume $\lambda$ is constant in this problem, and therefore the new intensity is
$
\nu =  \alpha \cdot \lambda.
$
Therefore, the maximum-reward motion problem in $\reals^2$, where a robot has agility $\alpha$ and the reward is generated by a Poisson point process with intensity $\lambda$, is equivalent to the maximum-reward motion problem with agility 1 and intensity $\alpha \lambda$, with respect to the mean reward collected.

Additionally, since the mean reward $\E[\totalreward (L)]$ grows proportionally with $\sqrt \lambda$, we conclude that it is also linear with $\sqrt \alpha$, \ie,
$$
\E[ \totalreward(L)] =  c\sqrt{\alpha} = c\sqrt{w/v}.
$$
for some constant $c>0$.
\end{proof}

%% file: main.bbl
\begin{thebibliography}{10}
\providecommand{\url}[1]{#1}
\csname url@samestyle\endcsname
\providecommand{\newblock}{\relax}
\providecommand{\bibinfo}[2]{#2}
\providecommand{\BIBentrySTDinterwordspacing}{\spaceskip=0pt\relax}
\providecommand{\BIBentryALTinterwordstretchfactor}{4}
\providecommand{\BIBentryALTinterwordspacing}{\spaceskip=\fontdimen2\font plus
\BIBentryALTinterwordstretchfactor\fontdimen3\font minus
  \fontdimen4\font\relax}
\providecommand{\BIBforeignlanguage}[2]{{%
\expandafter\ifx\csname l@#1\endcsname\relax
\typeout{** WARNING: IEEEtran.bst: No hyphenation pattern has been}%
\typeout{** loaded for the language `#1'. Using the pattern for}%
\typeout{** the default language instead.}%
\else
\language=\csname l@#1\endcsname
\fi
#2}}
\providecommand{\BIBdecl}{\relax}
\BIBdecl

\bibitem{bellmore1968traveling}
M.~Bellmore and G.~L. Nemhauser, ``The traveling salesman problem: a survey,''
  \emph{Operations Research}, vol.~16, no.~3, pp. 538--558, 1968.

\bibitem{lawler1985traveling}
E.~L. Lawler, ``The traveling salesman problem: a guided tour of combinatorial
  optimization,'' \emph{WILEY-INTERSCIENCE SERIES IN DISCRETE MATHEMATICS},
  1985.

\bibitem{laporte1992traveling}
G.~Laporte, ``The traveling salesman problem: An overview of exact and
  approximate algorithms,'' \emph{European Journal of Operational Research},
  vol.~59, no.~2, pp. 231--247, 1992.

\bibitem{applegate2011traveling}
D.~L. Applegate, R.~E. Bixby, V.~Chvatal, and W.~J. Cook, \emph{The Traveling
  Salesman Problem: A Computational Study}.\hskip 1em plus 0.5em minus
  0.4em\relax Princeton university press, 2011.

\bibitem{lin1973effective}
S.~Lin and B.~W. Kernighan, ``An effective heuristic algorithm for the
  traveling-salesman problem,'' \emph{Operations research}, vol.~21, no.~2, pp.
  498--516, 1973.

\bibitem{rosenkrantz1974approximate}
D.~J. Rosenkrantz, R.~E. Stearns, and P.~Lewis, ``Approximate algorithms for
  the traveling salesperson problem,'' in \emph{Switching and Automata Theory,
  1974., IEEE Conference Record of 15th Annual Symposium on}.\hskip 1em plus
  0.5em minus 0.4em\relax IEEE, 1974, pp. 33--42.

\bibitem{padberg1991branch}
M.~Padberg and G.~Rinaldi, ``A branch-and-cut algorithm for the resolution of
  large-scale symmetric traveling salesman problems,'' \emph{SIAM review},
  vol.~33, no.~1, pp. 60--100, 1991.

\bibitem{dorigo2014ant}
M.~Dorigo and L.~Gambardella, ``Ant-q: A reinforcement learning approach to the
  traveling salesman problem,'' in \emph{Proceedings of ML-95, Twelfth Intern.
  Conf. on Machine Learning}, 2014, pp. 252--260.

\bibitem{beardwood1959shortest}
J.~Beardwood, J.~H. Halton, and J.~M. Hammersley, ``The shortest path through
  many points,'' in \emph{Mathematical Proceedings of the Cambridge
  Philosophical Society}, vol.~55, no.~04.\hskip 1em plus 0.5em minus
  0.4em\relax Cambridge Univ Press, 1959, pp. 299--327.

\bibitem{savla2008traveling}
K.~Savla, E.~Frazzoli, and F.~Bullo, ``Traveling salesperson problems for the
  dubins vehicle,'' \emph{Automatic Control, IEEE Transactions on}, vol.~53,
  no.~6, pp. 1378--1391, 2008.

\bibitem{le2012dubins}
J.~Le~Ny, E.~Feron, and E.~Frazzoli, ``On the dubins traveling salesman
  problem.'' \emph{IEEE Trans. Automat. Contr.}, vol.~57, no.~1, pp. 265--270,
  2012.

\bibitem{dantzig1959truck}
G.~B. Dantzig and J.~H. Ramser, ``The truck dispatching problem,''
  \emph{Management science}, vol.~6, no.~1, pp. 80--91, 1959.

\bibitem{christofides1976vehicle}
N.~Christofides, ``The vehicle routing problem,'' \emph{Revue fran{\c{c}}aise
  d'automatique, d'informatique et de recherche op{\'e}rationnelle. Recherche
  op{\'e}rationnelle}, vol.~10, no.~1, pp. 55--70, 1976.

\bibitem{toth2001vehicle}
P.~Toth and D.~Vigo, \emph{The vehicle routing problem}.\hskip 1em plus 0.5em
  minus 0.4em\relax Society for Industrial and Applied Mathematics, 2001.

\bibitem{golden2008vehicle}
B.~L. Golden, S.~Raghavan, and E.~A. Wasil, \emph{The Vehicle Routing Problem:
  Latest Advances and New Challenges: latest advances and new
  challenges}.\hskip 1em plus 0.5em minus 0.4em\relax Springer Science \&
  Business Media, 2008, vol.~43.

\bibitem{laporte1992vehicle}
G.~Laporte, ``The vehicle routing problem: An overview of exact and approximate
  algorithms,'' \emph{European Journal of Operational Research}, vol.~59,
  no.~3, pp. 345--358, 1992.

\bibitem{desrochers1992new}
M.~Desrochers, J.~Desrosiers, and M.~Solomon, ``A new optimization algorithm
  for the vehicle routing problem with time windows,'' \emph{Operations
  research}, vol.~40, no.~2, pp. 342--354, 1992.

\bibitem{osman1993metastrategy}
I.~H. Osman, ``Metastrategy simulated annealing and tabu search algorithms for
  the vehicle routing problem,'' \emph{Annals of operations research}, vol.~41,
  no.~4, pp. 421--451, 1993.

\bibitem{gendreau1994tabu}
M.~Gendreau, A.~Hertz, and G.~Laporte, ``A tabu search heuristic for the
  vehicle routing problem,'' \emph{Management science}, vol.~40, no.~10, pp.
  1276--1290, 1994.

\bibitem{baldacci2012recent}
R.~Baldacci, A.~Mingozzi, and R.~Roberti, ``Recent exact algorithms for solving
  the vehicle routing problem under capacity and time window constraints,''
  \emph{European Journal of Operational Research}, vol. 218, no.~1, pp. 1--6,
  2012.

\bibitem{wilson1977computer}
N.~H. Wilson and N.~J. Colvin, \emph{Computer control of the Rochester
  dial-a-ride system}.\hskip 1em plus 0.5em minus 0.4em\relax Massachusetts
  Institute of Technology, Center for Transportation Studies, 1977.

\bibitem{gendreau1996stochastic}
M.~Gendreau, G.~Laporte, and R.~S{\'e}guin, ``Stochastic vehicle routing,''
  \emph{European Journal of Operational Research}, vol.~88, no.~1, pp. 3--12,
  1996.

\bibitem{bertsimas1991stochastic}
D.~J. Bertsimas and G.~Van~Ryzin, ``A stochastic and dynamic vehicle routing
  problem in the euclidean plane,'' \emph{Operations Research}, vol.~39, no.~4,
  pp. 601--615, 1991.

\bibitem{kataoka1988algorithm}
S.~Kataoka and S.~Morito, ``An algorithm for single constraint maximum
  collection problem.'' \emph{J. OPER. RES. SOC. JAPAN.}, vol.~31, no.~4, pp.
  515--530, 1988.

\bibitem{butt1994heuristic}
S.~E. Butt and T.~M. Cavalier, ``A heuristic for the multiple tour maximum
  collection problem,'' \emph{Computers \& Operations Research}, vol.~21,
  no.~1, pp. 101--111, 1994.

\bibitem{laporte1990selective}
G.~Laporte and S.~Martello, ``The selective travelling salesman problem,''
  \emph{Discrete applied mathematics}, vol.~26, no.~2, pp. 193--207, 1990.

\bibitem{feillet2005traveling}
D.~Feillet, P.~Dejax, and M.~Gendreau, ``Traveling salesman problems with
  profits,'' \emph{Transportation science}, vol.~39, no.~2, pp. 188--205, 2005.

\bibitem{arkin1998resource}
E.~M. Arkin, J.~S. Mitchell, and G.~Narasimhan, ``Resource-constrained
  geometric network optimization,'' in \emph{Proceedings of the fourteenth
  annual symposium on Computational geometry}.\hskip 1em plus 0.5em minus
  0.4em\relax ACM, 1998, pp. 307--316.

\bibitem{golden1987orienteering}
B.~L. Golden, L.~Levy, and R.~Vohra, ``The orienteering problem,'' \emph{Naval
  Research Logistics (NRL)}, vol.~34, no.~3, pp. 307--318, 1987.

\bibitem{vansteenwegen2011orienteering}
P.~Vansteenwegen, W.~Souffriau, and D.~Van~Oudheusden, ``The orienteering
  problem: A survey,'' \emph{European Journal of Operational Research}, vol.
  209, no.~1, pp. 1--10, 2011.

\bibitem{tsiligirides1984heuristic}
T.~Tsiligirides, ``Heuristic methods applied to orienteering,'' \emph{Journal
  of the Operational Research Society}, pp. 797--809, 1984.

\bibitem{chao1996fast}
I.-M. Chao, B.~L. Golden, and E.~A. Wasil, ``A fast and effective heuristic for
  the orienteering problem,'' \emph{European Journal of Operational Research},
  vol.~88, no.~3, pp. 475--489, 1996.

\bibitem{golden1988multifaceted}
B.~L. Golden, Q.~Wang, and L.~Liu, ``A multifaceted heuristic for the
  orienteering problem,'' \emph{Naval Research Logistics (NRL)}, vol.~35,
  no.~3, pp. 359--366, 1988.

\bibitem{fischetti1998solving}
M.~Fischetti, J.~J.~S. Gonzalez, and P.~Toth, ``Solving the orienteering
  problem through branch-and-cut,'' \emph{INFORMS Journal on Computing},
  vol.~10, no.~2, pp. 133--148, 1998.

\bibitem{leifer1994strong}
A.~C. Leifer and M.~B. Rosenwein, ``Strong linear programming relaxations for
  the orienteering problem,'' \emph{European Journal of Operational Research},
  vol.~73, no.~3, pp. 517--523, 1994.

\bibitem{geem2005harmony}
Z.~W. Geem, C.-L. Tseng, and Y.~Park, ``Harmony search for generalized
  orienteering problem: best touring in china,'' in \emph{Advances in natural
  computation}.\hskip 1em plus 0.5em minus 0.4em\relax Springer, 2005, pp.
  741--750.

\bibitem{chao1996team}
I.-M. Chao, B.~L. Golden, and E.~A. Wasil, ``The team orienteering problem,''
  \emph{European journal of operational research}, vol.~88, no.~3, pp.
  464--474, 1996.

\bibitem{tang2005tabu}
H.~Tang and E.~Miller-Hooks, ``A tabu search heuristic for the team
  orienteering problem,'' \emph{Computers \& Operations Research}, vol.~32,
  no.~6, pp. 1379--1407, 2005.

\bibitem{archetti2007metaheuristics}
C.~Archetti, A.~Hertz, and M.~G. Speranza, ``Metaheuristics for the team
  orienteering problem,'' \emph{Journal of Heuristics}, vol.~13, no.~1, pp.
  49--76, 2007.

\bibitem{vansteenwegen2009iterated}
P.~Vansteenwegen, W.~Souffriau, G.~V. Berghe, and D.~Van~Oudheusden, ``Iterated
  local search for the team orienteering problem with time windows,''
  \emph{Computers \& Operations Research}, vol.~36, no.~12, pp. 3281--3290,
  2009.

\bibitem{vansteenwegen2009guided}
------, ``A guided local search metaheuristic for the team orienteering
  problem,'' \emph{European Journal of Operational Research}, vol. 196, no.~1,
  pp. 118--127, 2009.

\bibitem{labadie2012team}
N.~Labadie, R.~Mansini, J.~Melechovsk{\`y}, and R.~W. Calvo, ``The team
  orienteering problem with time windows: An lp-based granular variable
  neighborhood search,'' \emph{European Journal of Operational Research}, vol.
  220, no.~1, pp. 15--27, 2012.

\bibitem{smith2011persistentAdapting}
R.~N. Smith, M.~Schwager, S.~L. Smith, B.~H. Jones, D.~Rus, and G.~S. Sukhatme,
  ``Persistent ocean monitoring with underwater gliders: Adapting sampling
  resolution,'' \emph{Journal of Field Robotics}, vol.~28, no.~5, pp. 714--741,
  2011.

\bibitem{garg2014persistent}
S.~Garg and N.~Ayanian, ``Persistent monitoring of stochastic spatio-temporal
  phenomena with a small team of robots,'' \emph{Proceedings of Robotics:
  Science and Systems, Berkeley, USA}, 2014.

\bibitem{cassandras2011optimal}
C.~G. Cassandras, X.~C. Ding, and X.~Lin, ``An optimal control approach for the
  persistent monitoring problem,'' in \emph{Decision and Control and European
  Control Conference (CDC-ECC), 2011 50th IEEE Conference on}.\hskip 1em plus
  0.5em minus 0.4em\relax IEEE, 2011, pp. 2907--2912.

\bibitem{lin2013optimal}
X.~Lin and C.~Cassandras, ``An optimal control approach to the multi-agent
  persistent monitoring problem in two-dimensional spaces,'' in \emph{Decision
  and Control (CDC), 2013 IEEE 52nd Annual Conference on}.\hskip 1em plus 0.5em
  minus 0.4em\relax IEEE, 2013, pp. 6886--6891.

\bibitem{cassandras2013optimal}
C.~G. Cassandras, X.~Lin, and X.~Ding, ``An optimal control approach to the
  multi-agent persistent monitoring problem,'' \emph{Automatic Control, IEEE
  Transactions on}, vol.~58, no.~4, pp. 947--961, 2013.

\bibitem{smith2011persistent}
S.~L. Smith, M.~Schwager, and D.~Rus, ``Persistent monitoring of changing
  environments using a robot with limited range sensing,'' in \emph{Robotics
  and Automation (ICRA), 2011 IEEE International Conference on}.\hskip 1em plus
  0.5em minus 0.4em\relax IEEE, 2011, pp. 5448--5455.

\bibitem{smith2012persistent}
------, ``Persistent robotic tasks: Monitoring and sweeping in changing
  environments,'' \emph{Robotics, IEEE Transactions on}, vol.~28, no.~2, pp.
  410--426, 2012.

\bibitem{alamdari2014persistent}
S.~Alamdari, E.~Fata, and S.~L. Smith, ``Persistent monitoring in discrete
  environments: Minimizing the maximum weighted latency between observations,''
  \emph{The International Journal of Robotics Research}, vol.~33, no.~1, pp.
  138--154, 2014.

\bibitem{alamdari2013min}
------, ``Min-max latency walks: Approximation algorithms for monitoring
  vertex-weighted graphs,'' in \emph{Algorithmic Foundations of Robotics
  X}.\hskip 1em plus 0.5em minus 0.4em\relax Springer, 2013, pp. 139--155.

\bibitem{yu2014persistent}
J.~Yu, S.~Karaman, and D.~Rus, ``Persistent monitoring of events with
  stochastic arrivals at multiple stations,'' in \emph{Robotics and Automation
  (ICRA), 2014 IEEE International Conference on}.\hskip 1em plus 0.5em minus
  0.4em\relax IEEE, 2014, pp. 5758--5765.

\bibitem{smith2009dynamic}
S.~L. Smith, S.~D. Bopardikar, and F.~Bullo, ``A dynamic boundary guarding
  problem with translating targets,'' in \emph{Decision and Control, 2009 held
  jointly with the 2009 28th Chinese Control Conference. CDC/CCC 2009.
  Proceedings of the 48th IEEE Conference on}.\hskip 1em plus 0.5em minus
  0.4em\relax IEEE, 2009, pp. 8543--8548.

\bibitem{bopardikar2010dynamic}
S.~D. Bopardikar, S.~L. Smith, F.~Bullo, and J.~P. Hespanha, ``Dynamic vehicle
  routing for translating demands: Stability analysis and receding-horizon
  policies,'' \emph{Automatic Control, IEEE Transactions on}, vol.~55, no.~11,
  pp. 2554--2569, 2010.

\bibitem{matni:design05}
N.~Matni and V.~Chandrasekaran, ``Regularization for design,'' California
  Institute of Technology, Tech. Rep., 2015.

\bibitem{censi16codesign}
A.~Censi, ``A mathematical theory of co-design,'' Laboratory for Information
  and Decision Systems/MIT, Tech. Rep., January 2016.

\bibitem{censi2017uncertainty}
------, ``Uncertainty in monotone co-design problems,'' \emph{IEEE Robotics and
  Automation Letters}, 2017.

\bibitem{karaman2012high}
S.~Karaman and E.~Frazzoli, ``High-speed flight in an ergodic forest,'' in
  \emph{Robotics and Automation (ICRA), 2012 IEEE International Conference
  on}.\hskip 1em plus 0.5em minus 0.4em\relax IEEE, 2012, pp. 2899--2906.

\bibitem{karaman2012highcdc}
------, ``High-speed motion with limited sensing range in a poisson forest,''
  in \emph{2012 IEEE 51st IEEE Conference on Decision and Control (CDC)}.\hskip
  1em plus 0.5em minus 0.4em\relax IEEE, 2012, pp. 3735--3740.

\bibitem{choudhurytheoretical}
S.~Choudhury, S.~Scherer, and J.~A. Bagnell, ``Theoretical limits of speed and
  resolution for kinodynamic planning in a poisson forest.''

\bibitem{rolla2008last}
L.~T. Rolla and A.~Q. Teixeira, ``Last passage percolation in macroscopi-cally
  inhomogeneous media,'' \emph{Electronic Communications in Probability},
  vol.~13, pp. 131--139, 2008.

\bibitem{seppalainen1997increasing}
T.~Sepp{\"a}l{\"a}inen \emph{et~al.}, ``Increasing sequences of independent
  points on the planar lattice,'' \emph{The Annals of Applied Probability},
  vol.~7, no.~4, pp. 886--898, 1997.

\bibitem{seppalainen2009lecture}
T.~Sepp{\"a}l{\"a}inen, ``Lecture notes on the corner growth model,''
  \emph{Unpublished notes}, 2009.

\bibitem{zeng2013directed}
X.~Zeng, Z.~Hou, C.~Guo, and Y.~Guo, ``Directed last-passage percolation and
  random matrices,'' \emph{Advances in Mathematics}, vol.~42, no.~3, p.~3,
  2013.

\bibitem{Grimmet:2002vk}
G.~Grimmet and P.~Heimer, ``{Directed Percolation and Random Walk},'' in
  \emph{In and out of equilibrium: probability with a physics flavor}.\hskip
  1em plus 0.5em minus 0.4em\relax Birkhauser, 2002.

\bibitem{Hambly:2007wm}
B.~Hambly and J.~B. Martin, ``{Heavy tails in last-passage percolation},''
  \emph{Probability Theory and Related Fields}, 2007.

\bibitem{baccelli2000asymptotic}
F.~Baccelli, A.~Borovkov, and J.~Mairesse, ``Asymptotic results on infinite
  tandem queueing networks,'' \emph{Probability theory and related fields},
  vol. 118, no.~3, pp. 365--405, 2000.

\bibitem{glynn1991departures}
P.~W. Glynn and W.~Whitt, ``Departures from many queues in series,'' \emph{The
  Annals of Applied Probability}, pp. 546--572, 1991.

\bibitem{martin2006last}
J.~B. Martin, ``Last-passage percolation with general weight distribution,''
  \emph{Markov Process. Related Fields}, vol.~12, no.~2, pp. 273--299, 2006.

\bibitem{somanath2014controlling}
A.~Somanath, S.~Karaman, and K.~Youcef-Toumi, ``Controlling stochastic growth
  processes on lattices: Wildfire management with robotic fire extinguishers,''
  in \emph{53rd IEEE Conference on Decision and Control}.\hskip 1em plus 0.5em
  minus 0.4em\relax IEEE, 2014, pp. 1432--1437.

\bibitem{ma2015maximum}
F.~Ma and S.~Karaman, ``Maximum-reward motion in a stochastic environment: The
  nonequilibrium statistical mechanics perspective,'' in \emph{Algorithmic
  Foundations of Robotics XI}.\hskip 1em plus 0.5em minus 0.4em\relax Springer,
  2015, pp. 389--406.

\bibitem{ivancevic2007applied}
V.~G. I. T.~T. Ivancevic, \emph{Applied differential geometry: a modern
  introduction}.\hskip 1em plus 0.5em minus 0.4em\relax World Scientific, 2007.

\bibitem{daley2007introduction}
D.~J. Daley and D.~Vere-Jones, \emph{An introduction to the theory of point
  processes: volume II: general theory and structure}.\hskip 1em plus 0.5em
  minus 0.4em\relax Springer Science \& Business Media, 2007, vol.~2.

\bibitem{chiu2013stochastic}
S.~N. Chiu, D.~Stoyan, W.~S. Kendall, and J.~Mecke, \emph{Stochastic geometry
  and its applications}.\hskip 1em plus 0.5em minus 0.4em\relax John Wiley \&
  Sons, 2013.

\bibitem{kushner1971introduction}
H.~J. Kushner, \emph{Introduction to stochastic control}.\hskip 1em plus 0.5em
  minus 0.4em\relax Holt, Rinehart and Winston New York, 1971.

\bibitem{Urmson:2008fw}
C.~U. et~al., ``{Autonomous driving in urban environments: Boss and the Urban
  Challenge},'' \emph{Journal of Field Robotics}, vol.~25, no.~8, pp. 425--466,
  2008.

\bibitem{Koenig:2004ut}
S.~Koenig, M.~Likhachev, and D.~Furcy, ``{Lifelong planning
  A{\textasteriskcentered}},'' \emph{Artificial Intelligence}, 2004.

\bibitem{pivtoraiko2011kinodynamic}
M.~Pivtoraiko and A.~Kelly, ``Kinodynamic motion planning with state lattice
  motion primitives,'' in \emph{Intelligent Robots and Systems (IROS), 2011
  IEEE/RSJ International Conference on}.\hskip 1em plus 0.5em minus 0.4em\relax
  IEEE, 2011, pp. 2172--2179.

\bibitem{cirillo2014lattice}
M.~Cirillo, T.~Uras, and S.~Koenig, ``A lattice-based approach to multi-robot
  motion planning for non-holonomic vehicles,'' in \emph{Intelligent Robots and
  Systems (IROS 2014), 2014 IEEE/RSJ International Conference on}.\hskip 1em
  plus 0.5em minus 0.4em\relax IEEE, 2014, pp. 232--239.

\bibitem{Schrijver:2003tv}
A.~Schrijver, \emph{{Combinatorial optimization}}.\hskip 1em plus 0.5em minus
  0.4em\relax Springer Verlag, 2003.

\bibitem{rolski2009stochastic}
T.~Rolski, H.~Schmidli, V.~Schmidt, and J.~Teugels, \emph{Stochastic processes
  for insurance and finance}.\hskip 1em plus 0.5em minus 0.4em\relax John Wiley
  \& Sons, 2009, vol. 505.

\bibitem{Angel2012}
\BIBentryALTinterwordspacing
O.~Angel and A.~Tomberg, ``Last passage percolation,'' \emph{Mprime Summer
  School in Probability}, 2012. [Online]. Available:
  \url{http://www.math.ubc.ca/~angel/ssprob12/courses.php}
\BIBentrySTDinterwordspacing

\bibitem{cator2011hydrodynamical}
E.~Cator and L.~P. Pimentel, ``Hydrodynamical methods in last passage
  percolation models,'' \emph{arXiv preprint arXiv:1106.2687}, 2011.

\bibitem{Steele:1996wn}
M.~J. Steele, \emph{{Probability Theory and Combinatorial Optimization}}.\hskip
  1em plus 0.5em minus 0.4em\relax SIAM, 1996.

\bibitem{johansson2001random}
K.~Johansson, ``Random growth and random matrices,'' in \emph{European Congress
  of Mathematics}.\hskip 1em plus 0.5em minus 0.4em\relax Springer, 2001, pp.
  445--456.

\bibitem{hambly2007heavy}
B.~Hambly and J.~B. Martin, ``Heavy tails in last-passage percolation,''
  \emph{probability theory and related fields}, vol. 137, no. 1-2, pp.
  227--275, 2020.

\bibitem{billingsley2009convergence}
P.~Billingsley, \emph{Convergence of probability measures}.\hskip 1em plus
  0.5em minus 0.4em\relax John Wiley \& Sons, 2009, vol. 493.

\bibitem{royden1988real}
H.~L. Royden and P.~Fitzpatrick, \emph{Real analysis}.\hskip 1em plus 0.5em
  minus 0.4em\relax Macmillan New York, 1988, vol. 198, no.~8.

\bibitem{martin2002linear}
J.~B. Martin, ``Linear growth for greedy lattice animals,'' \emph{Stochastic
  Processes and their Applications}, vol.~98, no.~1, pp. 43--66, 2002.

\bibitem{talagrand1995concentration}
M.~Talagrand, ``Concentration of measure and isoperimetric inequalities in
  product spaces,'' \emph{Publications Math{\'e}matiques de l'Institut des
  Hautes Etudes Scientifiques}, vol.~81, no.~1, pp. 73--205, 1995.

\bibitem{hille1996functional}
E.~Hille and R.~S. Phillips, \emph{Functional analysis and semi-groups}.\hskip
  1em plus 0.5em minus 0.4em\relax American Mathematical Soc., 1996, vol.~31.

\bibitem{streit2010Poisson}
R.~L. Streit, ``The poisson point process,'' in \emph{Poisson Point
  Processes}.\hskip 1em plus 0.5em minus 0.4em\relax Springer, 2010, pp.
  11--55.

\end{thebibliography}
